\documentclass[letterpaper]{article}
\pdfoutput=1
\usepackage{uai2020}
\usepackage[margin=1in]{geometry}
\usepackage{times}
\usepackage[numbers]{natbib}
\usepackage[colorlinks=true,citecolor=blue]{hyperref}
\usepackage{graphicx,xcolor,textcomp}
\graphicspath{{./Figures/}}
\usepackage{amsmath,amssymb,algorithm,bm}
\usepackage[noend]{algpseudocode}
\usepackage{enumitem}

\def\BibTeX{{\rm B\kern-.05em{\sc i\kern-.025em b}\kern-.08em
    T\kern-.1667em\lower.7ex\hbox{E}\kern-.125emX}}

\newcommand{\x}{\mathbf{x}}
\newcommand{\q}{\mathbf{q}}
\newcommand{\f}{{\bm{f}}}
\newcommand{\RR}{\mathbb{R}}
\newcommand{\NN}{\mathbb{N}}
\newcommand{\EE}{\mathbb{E}}
\newcommand{\GP}{{\cal GP}}
\newcommand{\wh}[1]{\widehat{#1}}
\newcommand{\wt}[1]{\widetilde{#1}}

\newcommand{\emem}[1]{{\bf\textsf{#1}}}
\newcommand{\newbo}{BDC}

\title{Using Distance Correlation for Efficient Bayesian Optimization}

\author{{\bf Takuya Kanazawa}
\\
Research and Development Group, Hitachi, Ltd., Kokubunji, Tokyo 185-8601, Japan
\\
Current address: Kobe Gakuin University, 
1-1-3 Minatojima, Chuo-ku, Kobe, Hyogo 650-8586, Japan
}

\begin{document}

\maketitle

\begin{abstract}
The need to collect data via expensive measurements of black-box functions is prevalent across science, engineering and medicine. As an example, hyperparameter tuning of a large AI model is critical to its predictive performance but is generally time-consuming and unwieldy. Bayesian optimization (BO) is a collection of methods that aim to address this issue by means of Bayesian statistical inference. In this work, we put forward a BO scheme named \newbo, which integrates BO with a statistical measure of association of two random variables called \emph{Distance Correlation}. {\newbo} balances exploration and exploitation automatically, and requires no manual hyperparameter tuning. We evaluate {\newbo} on a range of benchmark tests and observe that it performs on per with popular BO methods such as the expected improvement and max-value entropy search. We also apply {\newbo} to optimization of sequential integral observations of an unknown terrain and confirm its utility.
\end{abstract}

\section{Introduction}

In optimization problems arising in science and industry, we often face black-box functions that are expensive to evaluate. In clinical medicine, computed tomography (CT) scans enable us to investigate internal organs accurately. However, the number of CT scans must be minimized to avoid excessive radiation exposure of the patient body, and doctors wish to optimize the scan procedure to balance the harm of exposure and the benefit of high-resolution imaging. In engineering, mining companies strive to discover underground veins of valuable metals such as lithium with a limited budget for geological surveys, which calls for a methodology to nail down promising land locations as efficiently as possible. 

These are instances of sequential decision making problems. In contrast to static planning problems that can be solved with a long-term fixed plan that stipulates all later actions at the very beginning, sequential (dynamic) decision making problems allow us to gain more benefit by adaptively tuning later actions taking the outcomes of earlier actions into account. Such problems have been tackled with methods such as dynamic programming (DP) \cite{Bertsekas2017}, reinforcement learning (RL) \cite{SuttonBarto}, and Bayesian optimization (BO) \cite{Garnett2023book}. BO is a powerful toolbox with strong information-theoretical underpinning that enables to choose the next action near-optimally by utilizing the information gained from past observations \cite{Brochu2010tutorial,Shahriari2016,Frazier2018tutorial}. While DP is specifically suited to problems with finite discrete action/state spaces and (deep) RL generally suffers from high sample complexity and overfitting to training environments \cite{SuttonBarto,Li2017,Franois_Lavet_2018}, BO does not require any prior learning and can handle a variety of discrete or continuous action/state spaces flexibly.

In BO, during a cycle of actions and observations, the most recommended next action is determined by maximizing the so-called \emph{acquisition function}. In essence, different acquisition functions reflect multiple ways to balance \emph{exploration} and \emph{exploitation}---namely, the question of whether one should search nearby points of the current best solution or remote points that have not been deeply probed yet. Examples of popular acquisition functions include Probability of Improvement (PI) \cite{Kushner1964}, Expected Improvement (EI) \cite{Jones1998}, GP-UCB \cite{Srinivas2012}, GP-MI \cite{Contal2014}, and Max-value Entropy Search (MES) \cite{Wang2017}. Despite its simplicity, EI is known to be surprisingly competitive in various tasks \cite{Brochu2010tutorial,Shahriari2016,Frazier2018tutorial,Wang2017}. It is known that PI and EI can be interpolated continuously \cite{kanazawa2022ijcnn}. 

In this paper, a new scheme for BO that utilizes \emph{distance correlation} (DC) \cite{Szekely2007,Szekely2013,Rizzo2016} is proposed. DC is a nonparametric measure of correlations between two random variables in arbitrary dimensions. DC takes a value between 0 and 1, and vanishes if and only if the two variables are statistically independent. In this regard it is similar to mutual information (MI), but DC is generally easier and faster to compute from limited data than MI \cite{Szekely2007,Szekely2013}. As far as we know, DC has been seldom utilized in the literature of BO. In this work, a new algorithm referred to as Bayesian inference with Distance Correlation (\newbo) is proposed. In the first part of the paper, {\newbo} is employed to arrange sequential measurements of a black-box function. It is assumed that each observation has a finite resolution. This problem setup has been motivated by satellite imagery researches for agriculture \cite{rs12091357,Burke_2021,s21113758}%
\footnote{Related works based on deep RL can be found in \cite{Uzkent2020,Ayush2020}.}
wherein it is desired to gain information of the cropland with as few observations at as low resolutions as possible, to minimize remote sensing costs. More concretely, the optimization task we solve here is to decide \emph{where} to observe next at \emph{which} level of resolution. We show that {\newbo} is quite suitable for performing this task. In the second part of the paper, we consider the conventional setup of BO, i.e., search for the maximum of a given black-box function. The acquisition function of {\newbo} is formulated, and its favorable performance against baselines is demonstrated empirically across multiple benchmark problems. 

This paper is organized as follows. In Section~\ref{sc:25rwds} the backgrounds of BO and DC are summarized, and related work is reviewed. In Section~\ref{sc:fe} the remote sensing problem is studied using {\newbo}. In Section~\ref{sc:244dfs} global optimization is tackled and comparison between {\newbo} and alternative methods is made. We conclude in Section~\ref{sc:ewtefsd}. Some technical materials are relegated to \hyperref[sc:523rwe]{Appendix}.

\section{Background and related work\label{sc:25rwds}}

\subsection{Generalized measures of correlation}

In statistics, the Pearson correlation coefficient measures a \emph{linear} dependence between two random variables, while the Spearman's rank correlation coefficient is sensitive to a \emph{monotonic} relationship between two random variables \cite{Hastie_book2009}. As these classical information coefficients have obvious limitations, various nonparametric measures of statistical association that can detect and quantify nonlinear relationships have been proposed. Pivotal examples include the Maximal Information Coefficient \cite{Reshef2011}, the
Hilbert-Schmidt independence criterion \cite{HSIC2005,HSIC2007}, the Energy Statistics \cite{Szekely2013}, the Distance Correlation (DC) \cite{Szekely2007,Szekely2013,Rizzo2016}, and the Maximum Mean Discrepancy \cite{Gretton_NIPS2006}. They are deeply interrelated under certain conditions \cite{Sejdinovic2013}. As a main focus of the present paper, we will use DC to probe nonlinear correlations between two random vectors $X$ and $Y$. A virtue of DC is that it works even when the Euclidean dimensions of $X$ and $Y$ are different. DC has been used to study functional behaviors of deep neural networks \cite{Zhen2022}.

\subsection{Gaussian processes}

BO's major workhorse is the Gaussian Process (GP) regression, also known as kriging in geosciences, which not only allows interpolation of measured points but also provides the estimate of uncertainty for new predictions \cite{Garnett2023book,RW_GP_book}. Similar to the Gaussian distribution that defines a distribution over vectors, GP defines a distribution over functions. Given a mean function $m(\x)$ and the covariance function $k(\x,\x')$, sampled functions $f$ of GP satisfy
\begin{align}
  k(\x,\x') = \EE[(f(\x)-m(\x))(f(\x')-m(\x'))]\,,
\end{align}
which we write as $f(\x) \sim \GP(m(\x),k(\x,\x'))$. 
The prediction at a new point $\x_*$ obeys a Gaussian distribution. The mean takes the form
\begin{align}
  f(\x_*) =\sum_{i=1}^{N}\alpha_i k(\x_i,\x_*)
\end{align}
where $\alpha$'s are coefficients and $\{\x_i\}_{i=1}^{N}$ denote the set of observed points. The variance obeys a more complex form; see \cite{RW_GP_book} for more details. The choice of the covariance function $k$ is essential for all kernel-based methods including GP regression; the squared-exponential (Gaussian) kernel and the Mat\'{e}rn kernel are among the most popular ones, suitable for modeling smoothly varying functions.

The fact that GP is closed under linear operations such as differentiation and integration \cite{RW_GP_book} adds to its practical utility. GP with derivative information of the objective function, with applications to BO, has been studied in \cite{Solak2002nips,Osborne2009,Ahmed2016,
Wu2017nips,Wu2017arXiv,Eriksson2018nips}. The knowledge of gradients was shown to accelerate the convergence of optimization. GP with integral observations, or binned data, has also been studied, e.g., in \cite{Smith2018,Adelsberg2018,Law2018nips,
Hendriks2018,Purisha2019,Jidling2019,Hamelijnck2019nips,Tanaka2019nips,
Yousefi2019nips,Tanskanen2020,Longi2020} and GP has proven to be effective in modelling spatially or temporarily aggregated data.

\subsection{Sequential optimization}

A lingering problem in optimization is the trade-off between \emph{exploration} and \emph{exploitation} \cite{SuttonBarto}. BO schemes such as PI lean towards exploitation, as they highly weigh the chance of (possibly small) improvement over the current best solution. EI mitigates this issue by taking the magnitude of expected improvement into account. Recently acquisition functions of BO based on information theory have come into wide use. Entropy Search (ES) \cite{Villemonteix2009,Hennig2012} and Predictive Entropy Search (PES) \cite{HernandezLobato2014} prioritize a point that possesses maximal information about the arg-max of the objective function. However, their implementation is technically challenging and involves multiple steps of nontrivial approximations. These difficulties may be alleviated by searching for a point with maximal information about the max (\emph{not} arg-max) value of the objective function \cite{Hoffman2015,Wang2017}. This scheme goes under the name of Output-space Predictive Entropy Search (OPES) \cite{Hoffman2015} or Max-value Entropy Search (MES) \cite{Wang2017}. MES has been generalized to multi-fidelity \cite{Takeno2020,Moss2020} and multi-objective \cite{NEURIPS2019_82edc5c9,pmlr-v119-suzuki20a} settings. 

The term \emph{multi-fidelity} in optimization means that one has access to several correlated sources of information that differ in their observation costs and accuracy \cite{Forrester2007}. The goal of multi-fidelity BO is to optimize a function while minimizing the accumulated cost of observations. Strictly speaking, our study in Section~\ref{sc:fe} differs from this in four respects. First, we aim to evaluate a function globally (much like creating a map of a terrain), rather than finding just its optimum. Second, we assume that all modes of observations are of equal cost. Third, we focus on the use of DC, in contrast to \cite{Takeno2020,Moss2020}. Fourth, we incorporate \emph{integral observations} of a black-box function, which has not been considered before in the literature of multi-fidelity BO. 

It is worth noting that {\newbo} we propose is conceptually similar to OPES \cite{Hoffman2015} and MES \cite{Wang2017} in that the next query point depends on sample functions drawn from the posterior distribution of GP. However, while \cite{Hoffman2015,Wang2017} are guided by the reduction of uncertainty (entropy) of the max value of the objective function due to a new observation, our {\newbo} circumvents it by computing DC.

\section{Global probing of a black-box function\label{sc:fe}}

In this section we consider how best to make sequential \emph{integral} observations of a black-box function. It is assumed that the integral width (viz.~resolution) of an observation can be chosen at our disposal. For an illustration purpose, suppose one wants to make a contour map of an island. A conventional point measurement (i.e., a zero-width integral observation) is precise but only tells the elevation of the measured point, whereas a large-width observation tells us the average elevation of the whole island, which could be more helpful for contour mapping. A yet another didactic example is optimization of an advertising design. Suppose one wishes to improve an ad of a new product. Then it is quicker and more effective to begin by asking participants of a survey an easy-to-answer question such as a simple ``thumbs up/down'' choice, rather than engaging in an in-depth interview with each participant about the current ad design. Such rough low-resolution inquiries at an early stage of the project can facilitate streamlining the cycle of improvement and feedback at later stages.

Formally stated, the task is to formulate a policy according to which the (potentially most effective) locus and width of the next observation of a black-box function can be determined sequentially. In order to decide which policy works better, we need a criterion to rate the quality of a procedure. We will use the coefficient of determination $R^2$ \cite{Hastie_book2009} for this purpose; $R^2$ is always less than or equal to $1$, and higher is better.

\subsection{Algorithm}

The proposed algorithm {\newbo} for globally probing a real-valued function $f(\x)$ defined over a Euclidean domain $\chi$ is presented in Algorithm~\ref{alg:gpdc1}. The essential step is line 14, where the magnitude of DC is used to select the most preferable integral width of the next observation. 

Several remarks are in order.
\begin{algorithm}[tb]
\caption{~~{\newbo} for global probing \label{alg:gpdc1}}
\begin{algorithmic}[1]
	\Require $\x_0 \in\chi$: Initial query point 
	\newline\textcolor{white}{\;-}\quad 
	$w_0 \geq 0$: Width of the initial observation
	\newline\textcolor{white}{\;-}\quad 
	$\{\wh{\x}_n\}_{n=1:N}\subset \chi$: Representative points of $\chi$
	\newline\textcolor{white}{\;-}\quad 
	$M\in\NN$: Number of samples to be drawn from 
	\newline\textcolor{white}{\;-}\quad \qquad \quad ~
	\,posterior distribution of $f$
	\newline\textcolor{white}{\;-}\quad 
	$\{w_j\}_{j=1:J}\subset\RR_{\geq 0}$: Set of allowed widths of 
	\newline\textcolor{white}{\;-}\quad \qquad \quad ~\,
	integral observations 
	\State Observe $f$ at $\x_0$ with width $w_0$ and get $y_0$
	\State $D_0\gets \{(w_0, \x_0, y_0)\}$
	\For{$t=1,2,\dots$} 
	\State Optimize hyperparameters of GP model using $D_{t-1}$
	\State Compute mean and covariance of $f$ over $\{\wh{\x}_n\}$
	\newline \textcolor{white}{\;-}\quad 
	using $D_{t-1}$
	\State Draw $M$ samples of $f$ over $\{\wh{\x}_n\}$ from GP and obtain 
	\newline \textcolor{white}{-} \quad 
	$\{\f_m\}_{m=1:M}$ where $\f_m\in\RR^N$
	\State \emem{DC-list} $\gets \emptyset$
	\For{$j=1,2,\dots,J$}
	\State Compute the predictive variance $\sigma_j^2(\x)$ for an
	\newline \textcolor{white}{\;-} \qquad 
	integral observation of $f$ at $\forall\x\in\{\wh{\x}_n\}_{n=1:N}$
	\newline \textcolor{white}{\;-} \qquad 
	with width $w_j$
	\State $\wt{\x}_j \gets \mathrm{arg\,max}_{\x}~\sigma_j^2(\x)$
	\State Observe $M$ samples of $f$ (line 6) at $\x=\wt{\x}_j$  
	\newline \textcolor{white}{\;-} \qquad 
	with width $w_j$ and obtain $\{{f}_m^{(j)}\}_{m=1:M}\in\RR^M$\!\!\!
	\State Compute distance correlation between  
	\newline \textcolor{white}{\;-} \qquad 
	$\{\f_m\}_{m=1:M}$ and 
	$\{{f}_m^{(j)}\}_{m=1:M}$
	\State Define \emem{DC-list}$[j]:=$ DC$(\{\f_m\}, \{{f}_m^{(j)}\})$
	\EndFor 
	\State $\wt{j}\gets 
	\underset{1\leq j\leq J}{\mathrm{arg\,max}}$ 
	\emem{DC-list}$[j]$
	\State Observe $f$ at $\x=\wt{\x}_{\wt{j}}$ with width $w_{\wt{j}}$ and get $y_t$
	\State $D_t \gets D_{t-1} \cup \big\{\big(w_{\wt{j}}, \wt{\x}_{\wt{j}}, y_t \big)\big\}$
	\EndFor 
\end{algorithmic}
\end{algorithm}
\begin{itemize}[itemsep=0pt, topsep=0pt]
	\item GP with integral observations necessarily involves computation of integrated kernels of the form
	\begin{equation}
		\hspace{-16pt}
		k(\x, \q) \equiv \int_\chi \text{d}\x'~k(\x, \x') G(\x',\q)
	\end{equation}
	\vspace{-\baselineskip}
	\begin{equation}
		k(\q, \q') \equiv \int_\chi \text{d}\x \int_\chi \text{d}\x'~
		k(\x, \x') G(\x,\q)G(\x',\q')
		\label{eq:fewf989sdf}
	\end{equation}
	with a domain-specific weight function $G$. These integrals are numerically expensive. As a result, the iterative hyperparameter-optimization step in line 4 of Algorithm~\ref{alg:gpdc1} becomes a numerical bottleneck; we had to limit the number of iterations to 20. 
	\item In line 10 of Algorithm~\ref{alg:gpdc1}, the point of maximal uncertainty is selected, because the information gain from observing a Gaussian variable $\sim \mathcal{N}(\mu,\sigma^2)$ is $\log \sigma$ up to an irrelevant constant \cite{CT_book}.%
	\footnote{It has been proven that a myopic greedy policy for sequential information maximization is close to optimal under general conditions \cite{Golovin2011,Chen2015}.} 
	\item In line 6 of Algorithm~\ref{alg:gpdc1}, the samples of $f$ over $\{\wh{\x}_n\}_{n=1:N}$ are drawn from the posterior distribution of GP. This procedure requires Cholesky decomposition of the $N\times N$ covariance matrix for $f$, which may be prohibitively hard in high dimensions. For instance, in 10 dimensions, if a mesh of length 20 is taken along each dimension, $N$ becomes as large as $20^{10}$, which is far too large for numerics. We sidestep this problem by employing a sampling method of \cite{SSGP2010} and \cite[Section~2.1]{HernandezLobato2014} that utilizes the spectral density of kernel functions and random feature maps.
\end{itemize}

\subsection{Experiment on synthetic data}\label{sc:tsd}

\subsubsection{Integral observations}

\begin{figure}[tb]
	\centering
	\includegraphics[width=.8\columnwidth]{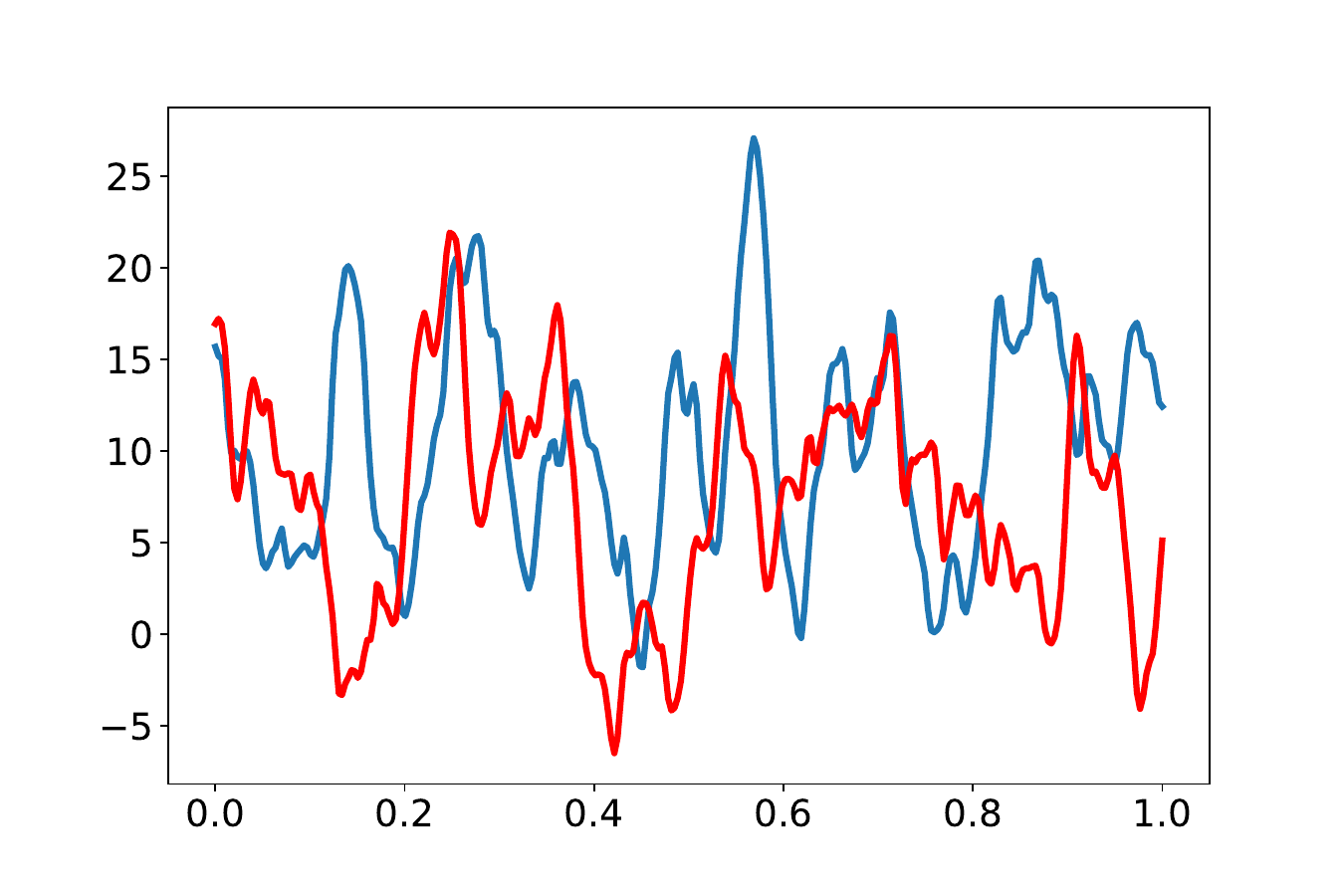}
	\caption{\label{fg:randcurve}Two samples of randomly generated multi-modal functions over the unit interval.}
\end{figure}

As a proof-of-concept experiment, we test the utility of {\newbo} as follows. We generated random continuous functions over the interval $[0,1]$, as shown in Figure~\ref{fg:randcurve}. They were sampled from GP endowed with the kernel function equal to the sum of the rational quadratic kernel and the Mat\'{e}rn 3/2 kernel, both of which have the length scale $0.02$ (we used {\tt GaussianProcessRegressor} implemented in the Python library {\tt scikit-learn} \cite{scikit-learn}). The sampled functions are highly multi-modal, as depicted in Figure~\ref{fg:randcurve}. It is assumed that $f(x)=f(0)$ for $x<0$ and $f(x)=f(1)$ for $x>1$. 

The problem setup is as follows. We sequentially perform 35 observations of a given function $f$. Each observation is either a point observation $f(x)$ ($w=0$), or an integral observation of width $w>0$:
\begin{align}
	\wt{f}_w(q):=\frac{1}{w}\int_{q-w/2}^{q+w/2}
	\!\!\! \text{d}x~f(x)\,, \quad 0\leq q\leq 1\,.
\end{align}
The two initial query points and their widths are chosen uniformly at random, and ensuing observations are coordinated according to Algorithm~\ref{alg:gpdc1}. As for a free parameter $0<\alpha<2$ of DC \cite{Szekely2007,Szekely2013}, we try $\alpha\in\{0.5, 1, 1.5\}$ to check $\alpha$-dependence of the effectiveness of Algorithm~\ref{alg:gpdc1}. $\{\wh{\x}_n\}_{n=1:N}$ are taken to be 120 equidistant points that uniformly cover $[0,1]$,  $M=200$, and $\{w_j\}_{j=1:J}=\{0, 0.0875, 0.175, 0.2625, 0.35, 0.4375, 0.525, \\
0.6125, 0.7\}$. Every time the mean of $f$ is updated (line 5) we evaluate the quality of our prediction by measuring $R^2$. We run Algorithm~\ref{alg:gpdc1} for $64$ different random functions and average results. 

As baselines, we have also implemented two other policies for sequential querying. The first is a random policy that selects both the next query point and the next width uniformly at random. The second is a zero-width max-variance policy, which avoids integral observations entirely and only performs point observations $(w=0)$ in such a way that the point that corresponds to the maximal predictive variance is selected as the next query point (as in line 10 of Algorithm~\ref{alg:gpdc1}). It makes no use of DC.

\begin{figure}[tb]
	\centering
	\includegraphics[width=.85\columnwidth]{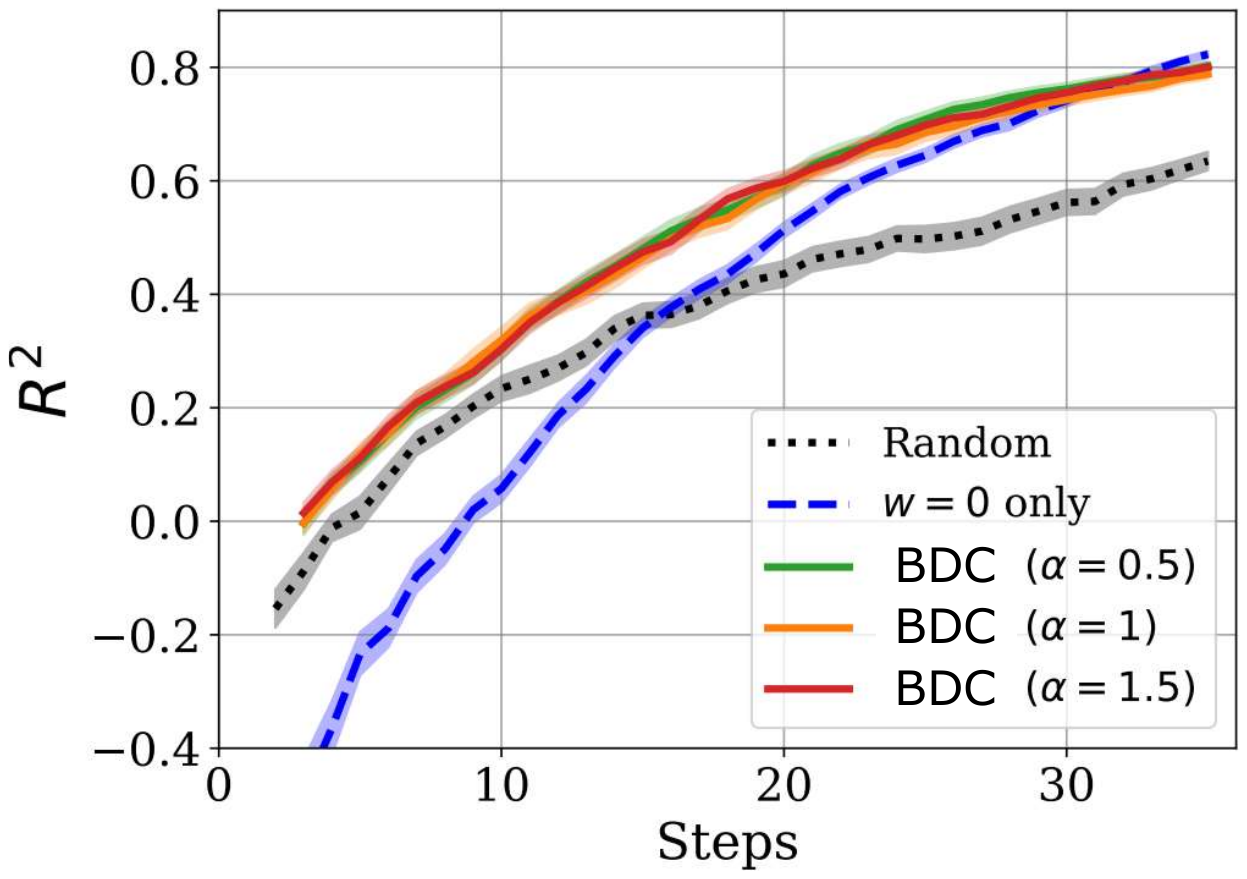}
	\caption{\label{fg:Mix_DC}Performance of {\newbo} and the two baseline methods. The error bands represent the standard deviation of the mean, i.e., the sample standard deviation divided by $\sqrt{64}=8$.}
	\vspace{\baselineskip}
	\includegraphics[width=\columnwidth]{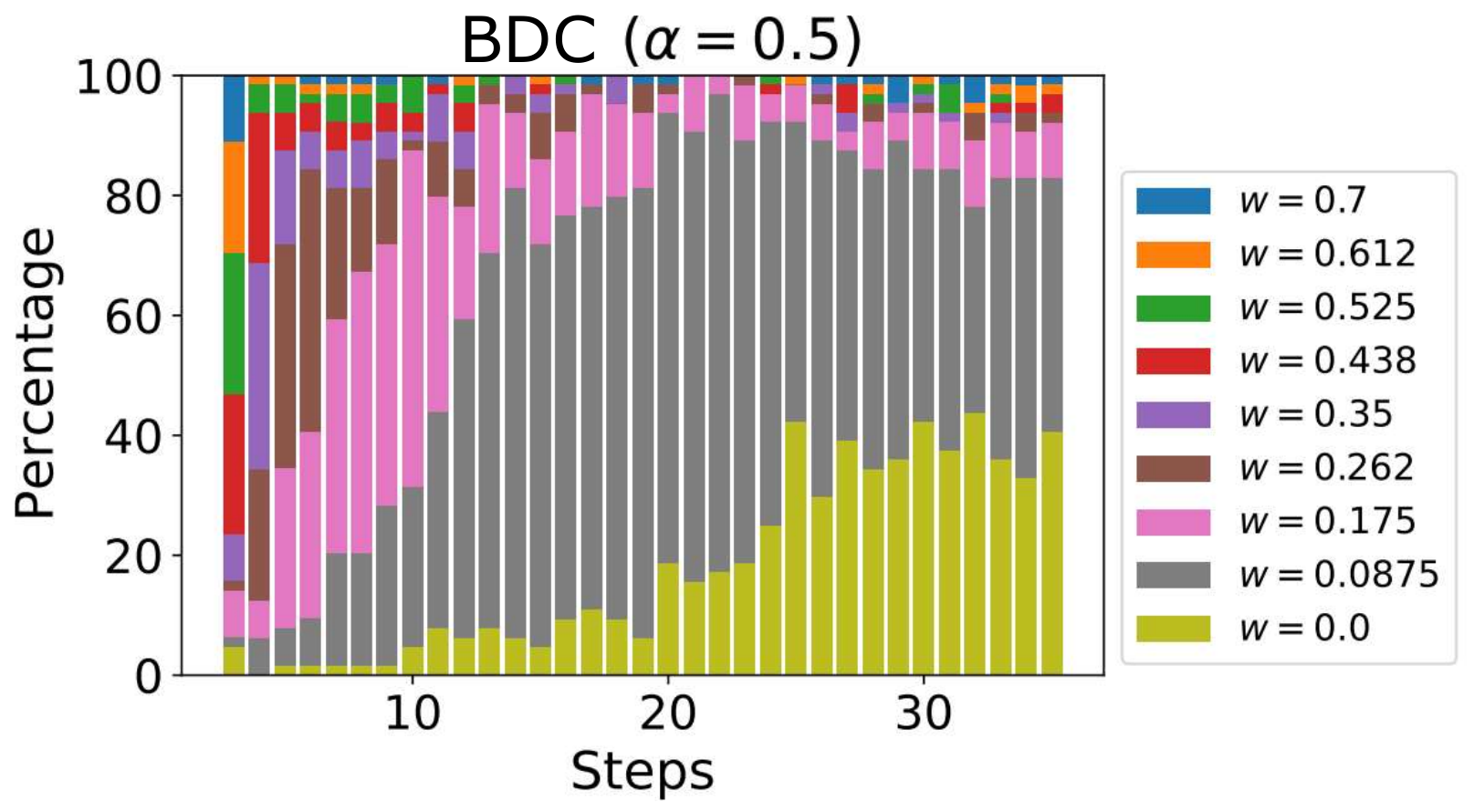}
	\caption{\label{fg:Mix_DC_w}Integral widths of iterative observations adaptively chosen by {\newbo} across 64 random functions.}
\end{figure}

The result of numerical experiments is shown in Figure~\ref{fg:Mix_DC}. The $\alpha$-dependence of {\newbo} appears to be very weak. {\newbo} is not only significantly better than the random policy but also outperforms the $w=0$ policy up to 32 steps. In fact, the latter exhibits a very low $R^2$ score for the initial steps due to the fact that point observations fail to capture the global terrain of a function. 

Figure~\ref{fg:Mix_DC_w} displays the distribution of observation widths adaptively chosen by {\newbo} ($\alpha=0.5$) at each step. It is noteworthy that {\newbo} favors relatively broad widths during the initial 10 steps, followed by an intermediate regime where the second smallest width $w=0.0875$ is dominant. Finally, a point observation gradually increases its proportion. This trend can be interpreted intuitively: broad observations are efficient for garnering the information on the global shape of the function, but it is soon exhausted and one has to resort to finer observations to acquire more local information. It is satisfactory to see that such adaptive decisions about the optimal resolution are made automatically by simply maximizing DC, with no subjective manual tuning of the policy. It is expected that the advantage of {\newbo} over the $w=0$ policy will diminish or even disappear if the objective function is dominated by long-wavelength Fourier modes, because local and global information are no longer sharply distinguished.

\subsubsection{Semi-local gradient observations}

The next question we would like to answer is whether {\newbo} can handle \emph{derivative (gradient) observations} or not. A gradient of a function carries quite local information and is complementary to non-local integral observations \cite{Wu2017nips}. Care must be taken in practice since a derivative observation is generally quite vulnerable to noises and numerical errors. To stabilize the optimization, we adopt a Gaussian filter of width $w>0$ before differentiation, leading us to define a \emph{semi-local gradient} as
\begin{align}
	\hspace{-4mm}
	\nabla^{[w]} f (\q) & \equiv \nabla_{\q} 
	\int_\chi \text{d}\x\; f(\x) \frac{1}{\sqrt{2\pi w^2}}\;
	\mathrm{e}^{-\parallel\x - \q \parallel^2/(2w^2)}
	\\
	& = \int_\chi \text{d}\x\; f(\x) \frac{1}{\sqrt{2\pi w^2}}\;
	\nabla_{\q}
	\mathrm{e}^{-\parallel\x - \q \parallel^2/(2w^2)}\,.
\end{align}
The test problem is set as follows. We first generate 48 random functions in the interval $[0,1]$. For each function $f$, $f(0)$ and $f(1)$ are observed first, and then 33 semi-local gradients are sequentially observed.  The width is selected from $\{w_j\}_{j=1:J}=$\{0.02, 0.06, 0.12, 0.2, 0.3, 0.4, 0.5, 0.6, 0.7\} according to {\newbo} $(\alpha=1)$. The number of posterior functions sampled for DC is $M=300$. For comparison, two other policies are also implemented: (i) a random policy that selects both the width and the query point $q\in[0,1]$ at random, and (ii) a narrow-gradient policy that exclusively adopts the smallest width $w=0.02$ and selects the point of maximal predictive variance sequentially.

\begin{figure}[tb]
	\centering
	\includegraphics[width=.85\columnwidth]{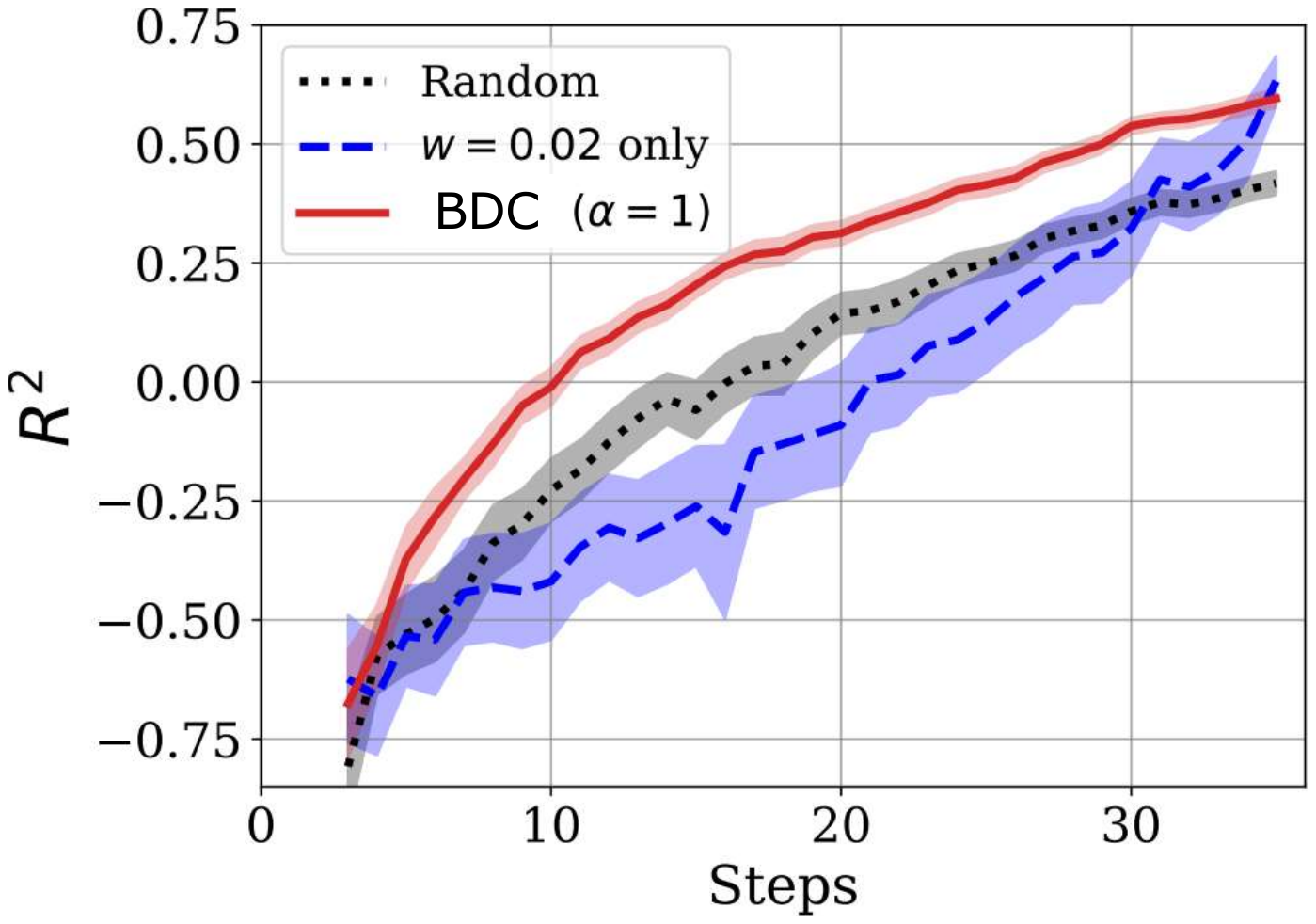}
	\caption{\label{fg:Deriv_DC}Performance of {\newbo} and the two baseline methods for semi-local gradient observations. The error bands represent the standard deviation of the mean, i.e., the sample standard deviation divided by $\sqrt{48}\simeq 6.93$.}
	\vspace{\baselineskip}
	\includegraphics[width=\columnwidth]{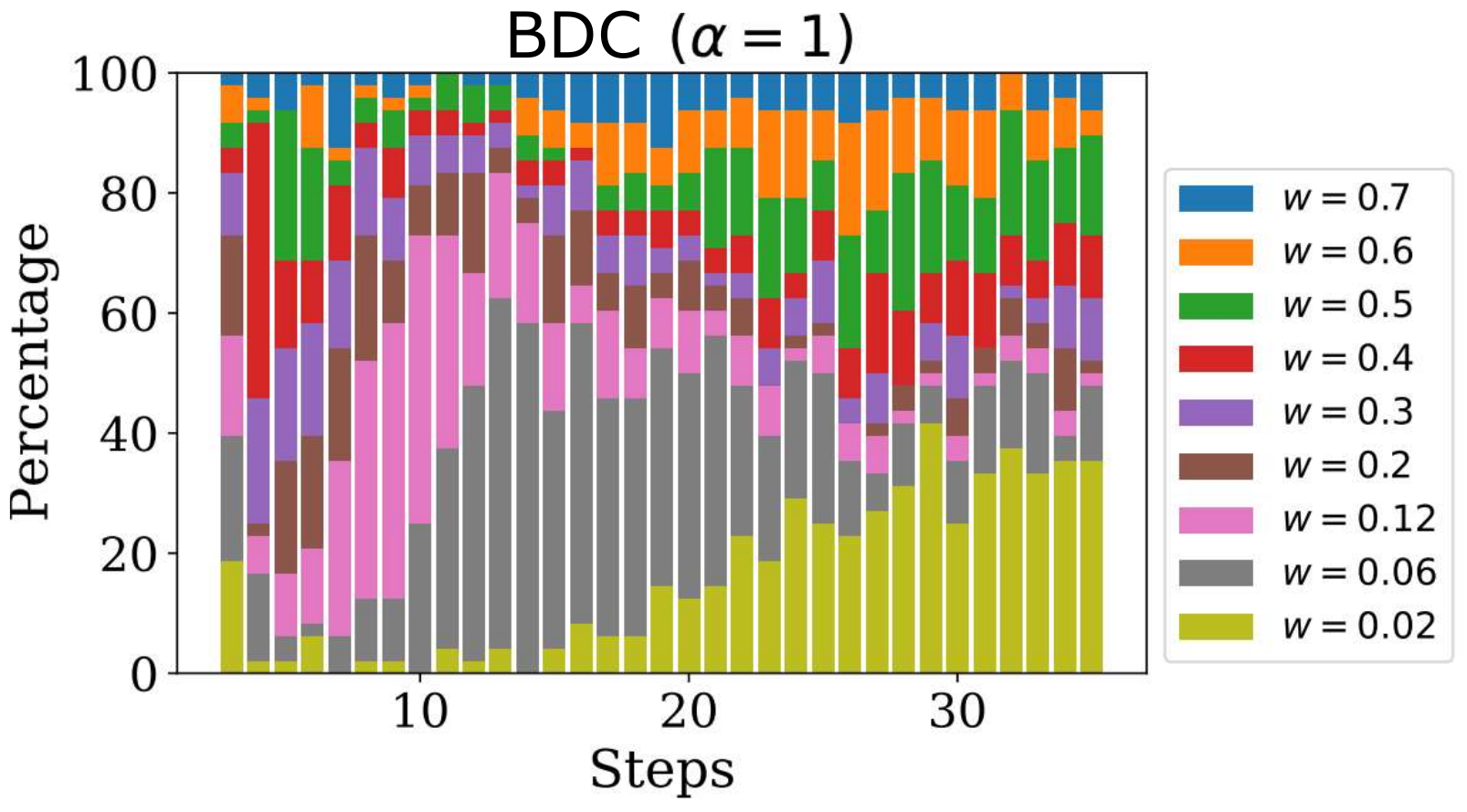}
	\caption{\label{fg:Deriv_DC_w}Widths of semi-local gradient observations adaptively chosen by {\newbo} for 48 random functions.}
\end{figure}

Figure~\ref{fg:Deriv_DC} summarizes the numerical results. Clearly {\newbo} shows superiority over the baselines. The $w=0.02$ policy catches up with {\newbo} at the 35th step, but its variance remains large over all steps. In contrast, the variance of {\newbo} is impressively small, indicating its stable performance for a variety of functions. In Figure~\ref{fg:Deriv_DC_w} we display the statistics of widths that were actually selected by {\newbo} at each step. One can see a monotonic increase of the proportion of the finest width $w=0.02$ as the iteration progresses, implying that local information assumes more importance after non-local information has been collected.

\subsection{Experiment on real data}

\subsubsection{Problem specification}

\begin{figure}[tb]
	\centering
	\includegraphics[width=.55\columnwidth]{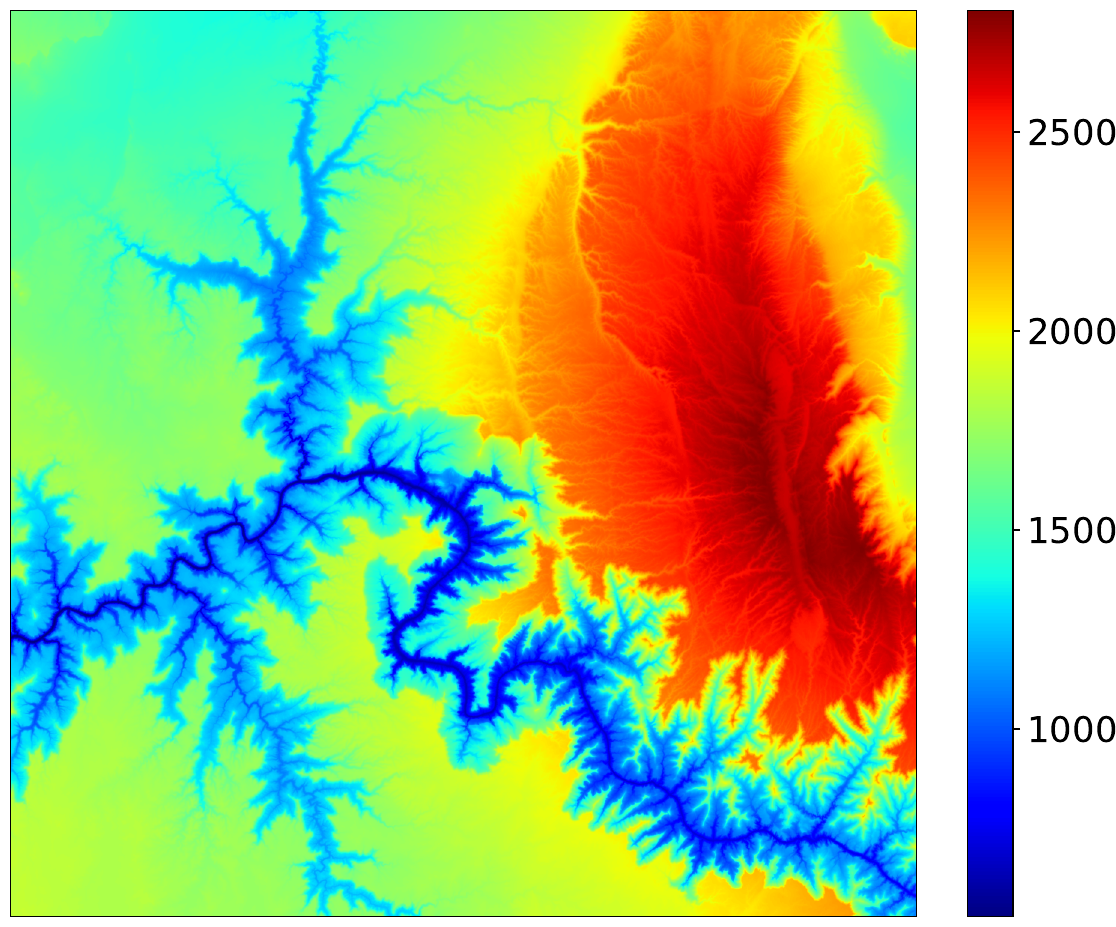}
	\caption{\label{fg:GC_image}Digital elevation model of the Grand Canyon in the United States \cite{USGS}.}
\end{figure}

We next apply the proposed algorithm to a remote sensing task of the landscape of the Grand Canyon, emulating surveillance from an airplane or a satellite. The digital elevation model available from \cite{USGS}, shown in Figure~\ref{fg:GC_image}, was used for this numerical experiment. The size of the image is $2800\times 2800$ pixels, which was rescaled to a unit square $[0,1]^2$. 
There are basically two approaches to implement integral observations in two dimensions: either use line integrals as in \cite{Purisha2019,Longi2020}, or use area integrals as in \cite{Tanaka2019nips}. The latter approach is adopted here. Specifically, we assume that the elevation of a terrain integrated over a disk of radius $r$ may be observed,
\begin{align}
    f_r(\q) \equiv \frac{1}{\pi r^2}
    \int_{\parallel \x-\q \parallel \leq r}
    \hspace{-2mm}\text{d}\x~f(\x)
    \label{eq:e53fsdf}
\end{align}
where $r\in\{0, 0.05, 0.1, 0.15, 0.2, 0.25, 0.3, 0.35, 0.4\}$. Note that $r=0$ is just the ordinary point observation. When the disk does not fit in $[0,1]^2$, $f$ is extended by reflection about the edges. We employ a quadrature formula in \cite[Theorem 1]{Bojanov1998} for efficient numerical integration over a disk, but \eqref{eq:e53fsdf} is still numerically expensive. The bottleneck is line 9 of Algorithm~\ref{alg:gpdc1}, where the variance needs to be computed for $N$ points simultaneously. It requires a computation of an $N\times N$ symmetric Gram matrix whose elements are double 2D (i.e., 4D) integrals of the kernel function (cf.~\eqref{eq:fewf989sdf}). Computation of these $N^2/2$ integrals becomes quickly infeasible when $N\gtrsim 100$. To cope with this issue, we make an evenly spaced meshgrid of size $30\times 30$ over $[0,1]^2$ and therefrom select 100 points randomly, for every $t$ and $j$ in Algorithm~\ref{alg:gpdc1}.

For comparison, two other policies are implemented: (i) a random policy that selects $\q$ and $r$ uniformly at random, and (ii) a policy that exclusively makes $r=0$ observations (i.e., a point with maximal predictive variance is sequentially observed). Each of the three algorithms is run for 16 different random seeds. We set $M=200$ for {\newbo}. 

\begin{figure}[tbh]
	\centering
	\includegraphics[width=.85\columnwidth]{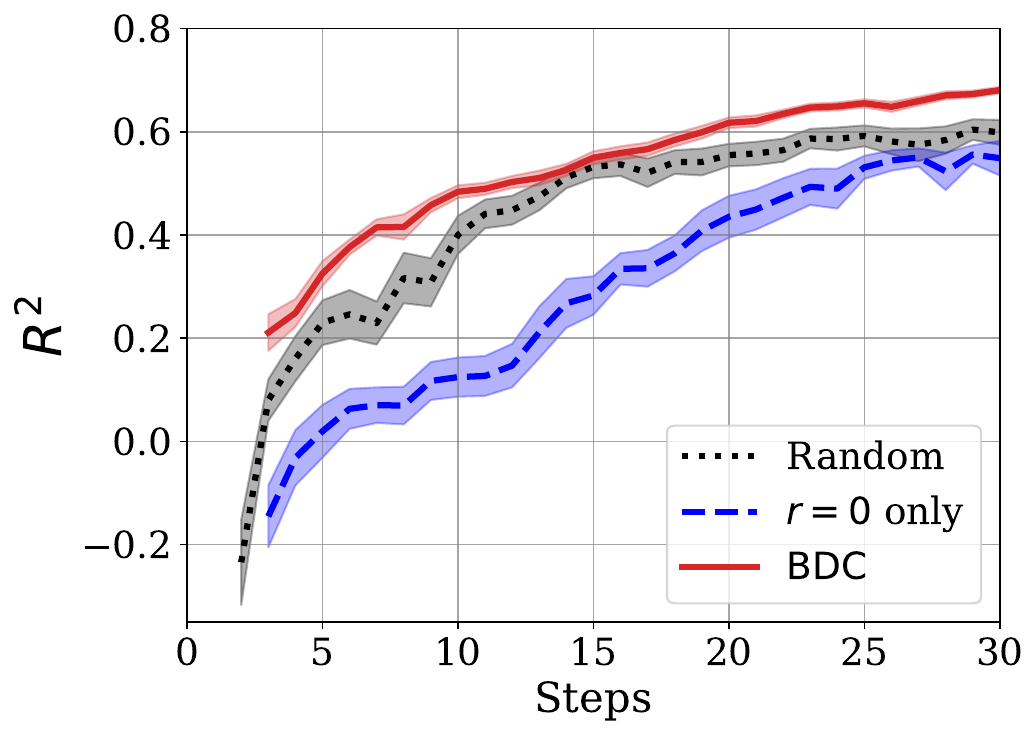}
	\caption{\label{fg:GC_DC}Performance of {\newbo} and the two baseline methods for evaluating the Grand Canyon data. The error bands represent the standard deviation of the mean, i.e., the sample standard deviation divided by $\sqrt{16}=4$.}
	\vspace{\baselineskip}
	\includegraphics[width=\columnwidth]{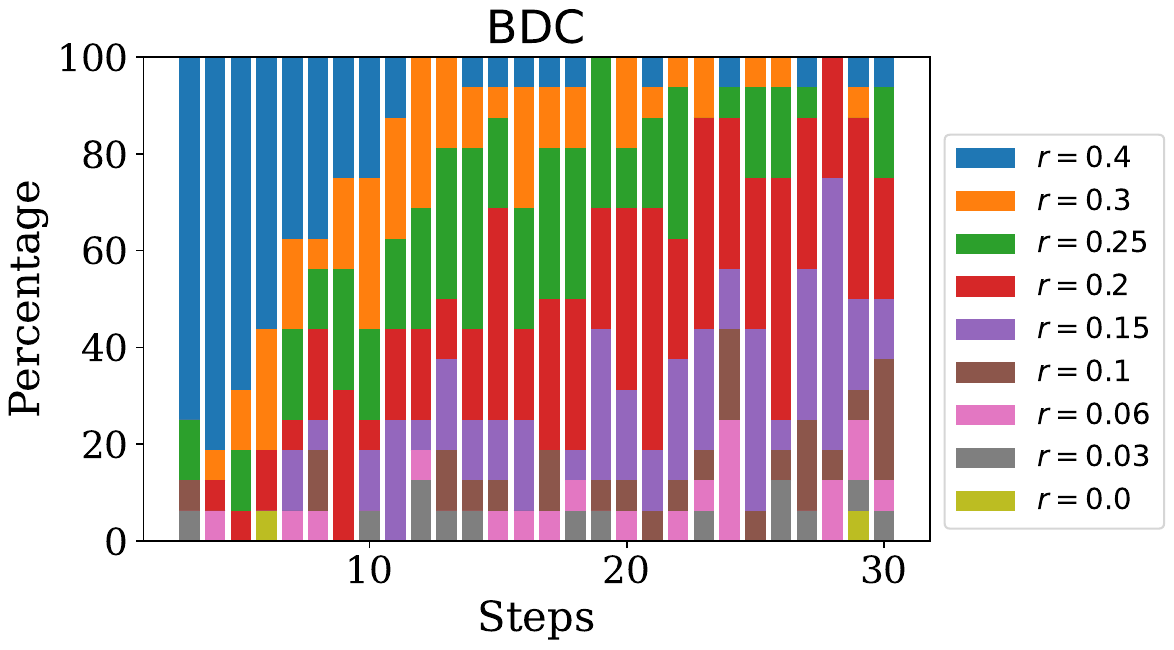}
	\caption{\label{fg:GC_DC_r}Widths of integral observations adaptively chosen by {\newbo} for 16 random initial conditions for evaluating the Grand Canyon data.}
\end{figure}

\subsubsection{Results}

The numerical results are presented in Figure~\ref{fg:GC_DC}. 
Clearly the policy with $r=0$ performs worst, indicating high efficiency of integral observations as a means of collecting macroscopic information. The gap between {\newbo} and the random policy is less impressive, which may be expected since the latter also performs integral observations albeit at random widths. The statistically significant difference between their performance, however, underlines the utility of {\newbo}, at least for this particular terrain case. 

The distribution of $r$ selected by {\newbo} at each step is shown in Figure~\ref{fg:GC_DC_r}. At the initial stages, $r\gtrsim0.3$ is strongly favored, whereas at the later stages $r\lesssim 0.15$ is favored. This figure brings to light that {\newbo} initially searches for long-wavelength information but later turns to shorter-wavelength information in a self-controlled manner, which in accordance with our findings in Section~\ref{sc:tsd}. 

The computation time of this experiment is summarized in Table~\ref{tb:243sd}. It is seen that the adoption of integral observations increases the numerical cost substantially. Whether this drawback is counterbalanced by the benefit of higher inference precision depends on specific use cases.

\begin{table}[tbh]
	\centering
	\caption{The wall-clock time of the numerical experiment in Figure~\ref{fg:GC_DC} for the three tested policies. We have used an E2 instance of Google's Vertex AI with 16 vCPUs, 8 Cores, and 128 GB of RAM. The runs with 16 random seeds were executed in parallel.\label{tb:243sd}\vspace{\baselineskip}}
	\begin{tabular}{c|c|c}
		Random policy & $r=0$ only & {\newbo}{\large \strut}
		\\\hline\hline
		5070 sec. & 1440 sec. & 44286 sec.{\large \strut}
	\end{tabular}
\end{table}

\section{Maximization of a black-box function \label{sc:244dfs}}

Next we turn to the standard BO setting where the optimum of a black-box function needs to be spotted with as few iterations of observation as possible \cite{Garnett2023book,Brochu2010tutorial,Shahriari2016,Frazier2018tutorial}.  
The term ``black-box'' implies that the explicit functional form of the objective is unknown, impeding the usage of white-box optimization approaches such as gradient-descent and Newton's method. A usual pitfall in this setting is to get stuck in a local optimum; thus any serious BO method must pursue a good compromise between exploration and exploitation. While the objective function $f$ can be defined in any domain including continuous, discrete, or hybrid spaces, we presume below that $f$'s domain is a subset of a Euclidean space so that continuous GP regression is applicable.

A common measure of performance of black-box optimization methods is the so-called \emph{regret}, which is the gap between the true optimum and the current best solution at hand. Formally it is defined at time step (viz.~iteration round) $T\geq 1$ as
\begin{equation}
	\text{Regret}_T \equiv \max_\x f(\x) - \max_{1\leq t \leq T} y_t 
	\label{eq:2d5dls}
\end{equation}
where $y_t$ is the actual observed value of $f$ at step $t$. Here we assume a noiseless observation. Regret is always non-negative and lower regret is a sign of better performance. It is desirable that regret decreases with $T$ as rapidly as possible.

\subsection{Algorithm}

Unlike in Section~\ref{sc:fe}, we shall limit ourselves to point observations of $f$ in the following, to reduce computational complexity. Our main proposal is Algorithm~\ref{alg:gpdc2}. The essential step is line 13, which determines the next query point by maximizing DC between the value of $\wh{f}(\x)$ and the global maximum value of $\wh{f}$ for every one of $M$ samples $(\ni \wh{f})$ drawn from the posterior distribution of GP. Note that even though the original objective $f$ may be expensive to evaluate, a posterior sample $\wh f$ is just a surrogate of $f$ and can be queried globally with little computational cost.

\begin{algorithm}[tb]
\caption{~~{\newbo} for maximum search \label{alg:gpdc2}}
\begin{algorithmic}[1]
	\Require $\x_0 \in\chi$: Initial query point 
	\newline\textcolor{white}{\;-}\quad 
	$\{\wh{\x}_n\}_{n=1:N}\subset \chi$: Representative points of $\chi$
	\newline\textcolor{white}{\;-}\quad 
	$M\in\NN$: Number of samples to be drawn from the
	\newline\textcolor{white}{\;-}\quad \qquad \quad ~\,
	posterior distribution of $f$
	\State Observe $f$ at $\x_0$ and get $y_0$
	\State $D_0 \gets \{(\x_0, y_0)\}$
	\For{$t=1,2,\dots$} 
	\State Optimize hyperparameters of GP model	using $D_{t-1}$
	\State Compute mean and covariance of $f$ over $\{\wh{\x}_n\}$
	\newline \textcolor{white}{\;-}\quad 
	using $D_{t-1}$
	\State Draw $M$ samples of $f$ over $\{\wh{\x}_n\}$ 
	\State Maximize each sample and obtain the set of max 
	\newline \textcolor{white}{\;-}\quad 
	values $\{{f}^{\max}_m\}_{m=1:M}\in\RR^M$
	\State \emem{DC-list} $\gets \emptyset$
	\For{$n=1,2,\dots,N$}
	\State For $M$ samples of $f$ in line 6, query the value 
	\newline \textcolor{white}{\;-} \qquad 
	at $\x=\wh{\x}_n$ and observe $\{{f}_m^{(n)}\}_{m=1:M}\in\RR^M$
	\State Compute DC between $\{{f}^{\max}_m\}_{m=1:M}$ and 
	\newline \textcolor{white}{\;-} \qquad 
	$\{{f}_m^{(n)}\}_{m=1:M}$
	\State Define \emem{DC-list}$[n]:=$ DC$(\{{f}^{\max}_m\}, \{{f}_m^{(n)}\})$
	\EndFor 
	\State $\wt{n} \gets 
	\underset{1\leq n \leq N}{\mathrm{arg\,max}}$ 
	\emem{DC-list}$[n]$
	\State Observe $f$ at $\x=\wh{\x}_{\wt{n}}$ and get $y_t$
	\State $D_t \gets D_{t-1} \cup \big\{\big(\wh{\x}_{\wt{n}}, y_t \big)\big\}$
	\EndFor 
\end{algorithmic}
\end{algorithm}

In line 7 of Algorithm~\ref{alg:gpdc2}, we could also collect arg\,max (rather than max) of each sample of $f$: $\{\x^{\max}_m\}_{m=1:M}\in\chi^M$. Line 12 should then be modified to \emem{DC-list}$[n]\equiv$ DC$\big(\{\x^{\max}_m\}, \{{f}_m^{(n)}\}\big)$. This variant will be called {\newbo-X}, to distinguish it from Algorithm~\ref{alg:gpdc2} that we refer to as {\newbo-y}.

\subsection{Experiment on synthetic data}

To check the efficacy of our algorithms we conduct numerical experiments adopting eight baselines: {Random}, {VarMax}, {PI}, {EI}, {GP-UCB}, {GP-MI}, {MES}, and {Brent}. 
Here, {VarMax} is a purely exploratory policy that queries the point of maximal predictive variance. For {PI} we use \cite[Eq.~(2)]{Brochu2010tutorial} with $\xi=10^{-3}$. For {GP-UCB} \cite{Srinivas2012}, we use \cite[Eq.~(5)]{Brochu2010tutorial} with $\nu=1$ and $\delta=0.05$. For {GP-MI} \cite{Contal2014}, we use $\alpha=\log(2/\delta)$ with $\delta=10^{-10}$. For {MES} we use \cite[Eq.~(6)]{Wang2017} with $K=300$. For {Brent} \cite{Brent1973}, we use the ${\tt minimize{\verb|_|}scalar}$ function of a Python library {\sf scipy} \cite{scipy}. Finally, we set $M=300$ for {\newbo} $(\alpha=1)$. 
\begin{figure}[!htbp]
	\centering
	\includegraphics[width=.85\columnwidth]{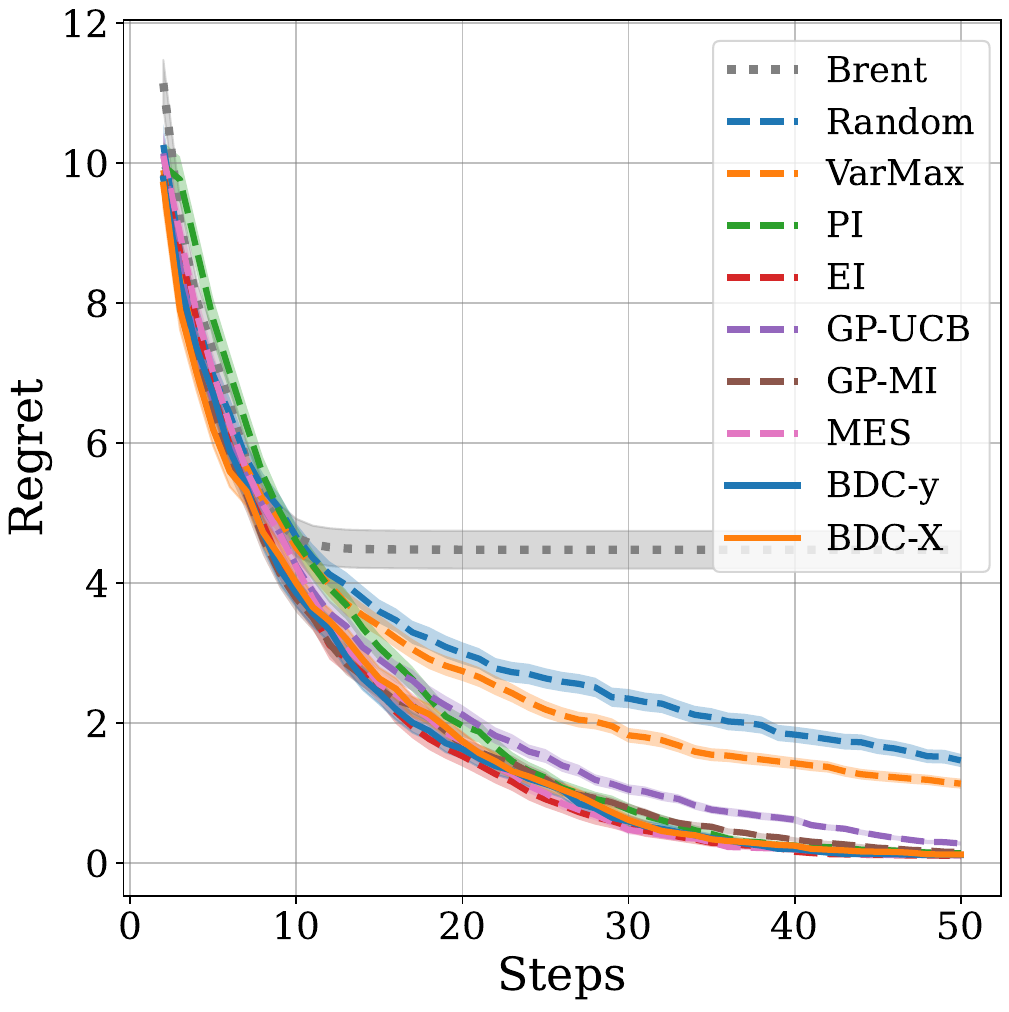}
	\caption{\label{fg:MaxTestAll}Performance of optimization algorithms evaluated on 256 random functions. The proposed algorithms are denoted by two solid lines. Error bands are one standard deviation of the mean.}
	\vspace{2\baselineskip}
	\includegraphics[width=.85\columnwidth]{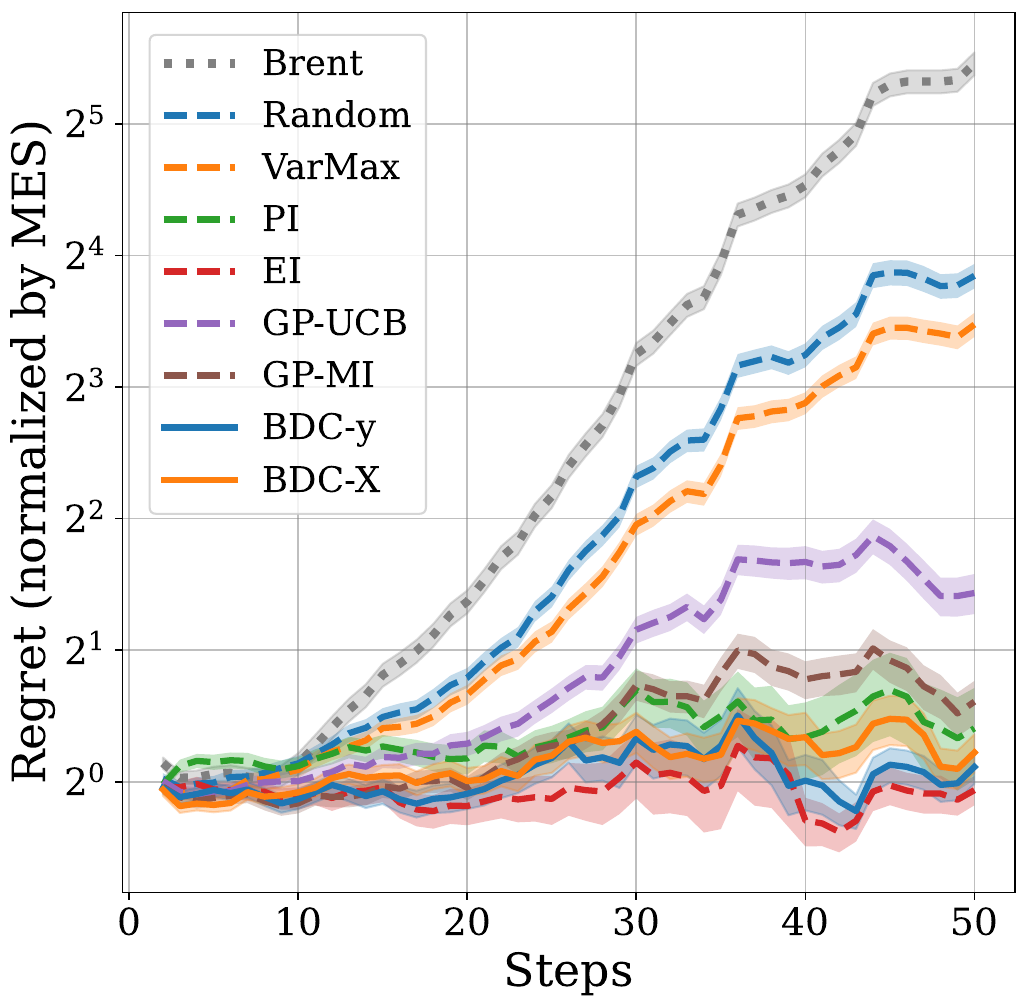}
	\caption{\label{fg:divided}Same as Figure~\ref{fg:MaxTestAll} but regret here is measured in unit of MES; namely, the regret at each step is divided by the mean regret of MES (MES becomes constant $1$ in this figure). Notice that the vertical axis is in log scale.}
\end{figure}

We generate 256 random functions over the unit interval $[0,1]$ in the manner of Section~\ref{sc:tsd}. For each function, the first two query points are chosen uniformly at random for all methods except for {Brent} for which all points including the first one are selected automatically. In total, 50 observations are made sequentially, and after every observation, the performance of each algorithm is measured by regret \eqref{eq:2d5dls}.

Figure~\ref{fg:MaxTestAll} displays average regret obtained from numerical simulations. Evidently the Brent method is stuck in a local optimum, while all the other algorithms seem to converge to the true optimum albeit at different speed. Random and VarMax converge significantly more slowly than others. 

For better visibility, we show in Figure~\ref{fg:divided} the regret \emph{normalized by that of MES} on log scale. Throughout the entire iterations, the regret of EI, {\newbo}-y and {\newbo}-X stay close to 1, implying that their performance is about the same as MES. The performance of GP-UCB seems to be slightly worse, especially at later iterations.

\begin{figure}[tb]
	\centering
	\includegraphics[width=.75\columnwidth]{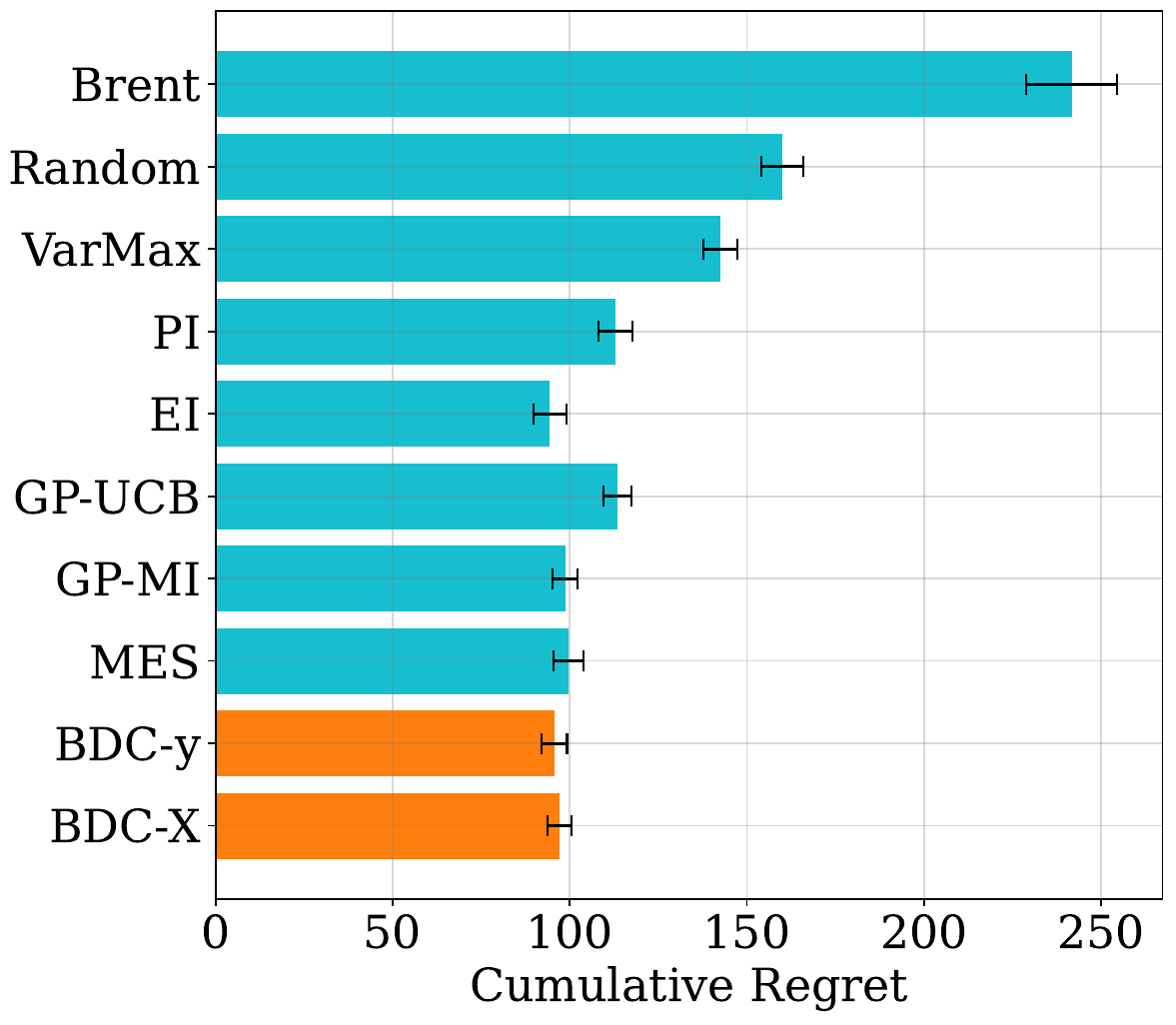}
	\caption{\label{fg:MaxTestAllBars}Cumulative regret of optimization algorithms averaged over 256 random functions (lower is better). The error bars denote one standard deviation of the mean.}
\end{figure}

To facilitate an intuitive and quantitative comparison, we show the \emph{cumulative regret} $\sum_{T=2}^{50}\text{Regret}_T$ in Figure~\ref{fg:MaxTestAllBars}. It is seen that the scores of EI, GP-MI, MES, {\newbo}-y and {\newbo}-X are nearly equal. We conducted the Wilcoxon signed-rank test \cite{Wilcoxon} and found no statistically significant difference among them. We conclude that our methods work as nicely as popular alternatives at least within this simple benchmark setting.

\subsection{Experiment on benchmark functions}\label{sc:bench}

Next we proceed to numerical tests on four popular benchmark functions in two dimensions: the Goldstein-Price function, Himmelblau function, Eggholder function, and Branin function. Details of these functions are described in \hyperref[sc:523rwe]{Appendix}. For each function, 50 observations are sequentially made, where the first two query points are chosen uniformly at random. We run each algorithm with 64 different random seeds and compare the averages. 

\begin{figure*}
    \centering
    \mbox{
    \includegraphics[width=.75\columnwidth]{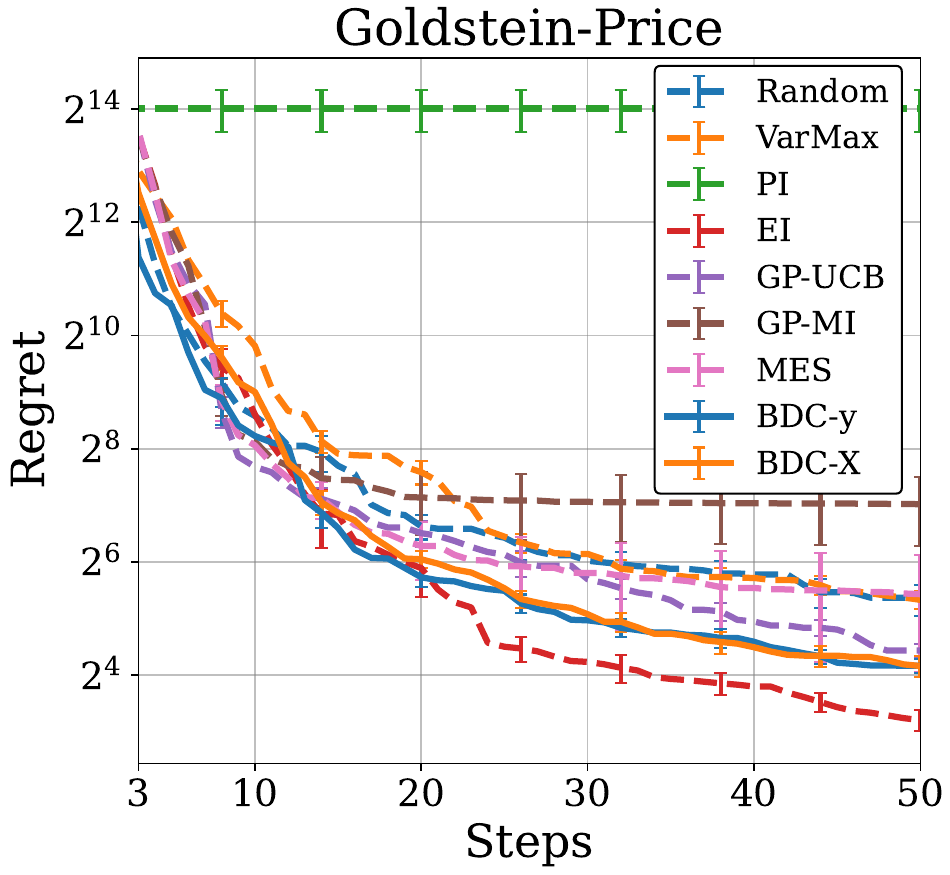}
    \quad
    \includegraphics[width=.75\columnwidth]{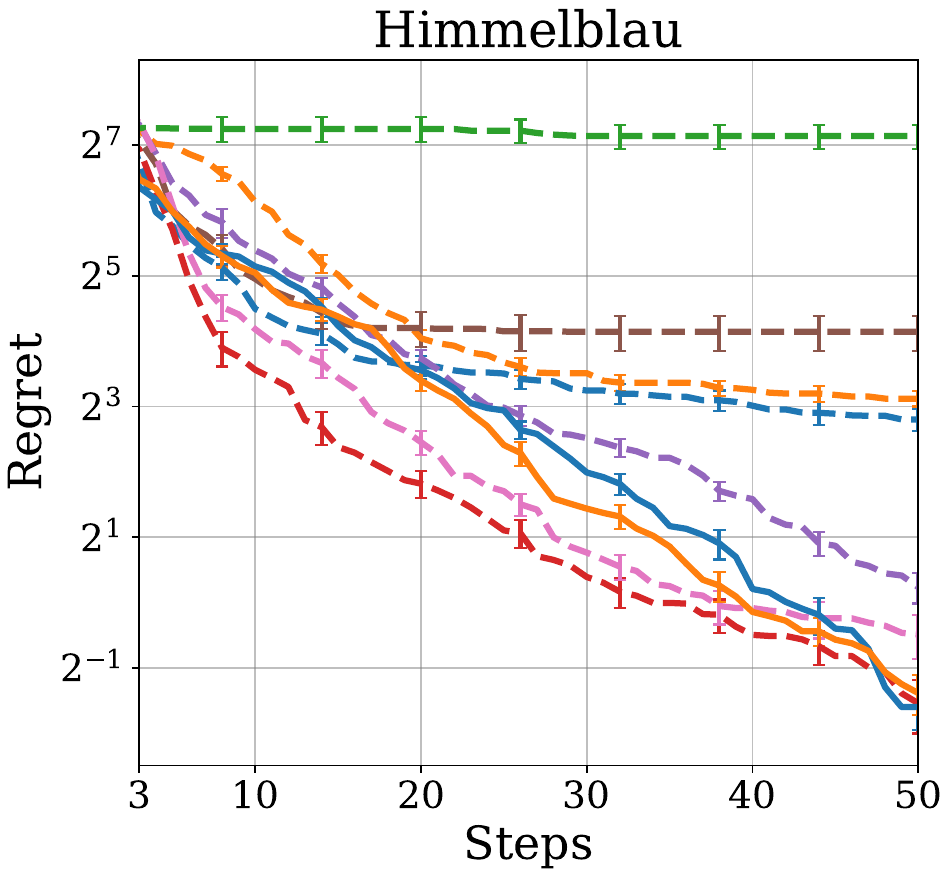}}
    \vspace{10pt}
    \\
    \centering
    \mbox{
    \includegraphics[width=.75\columnwidth]{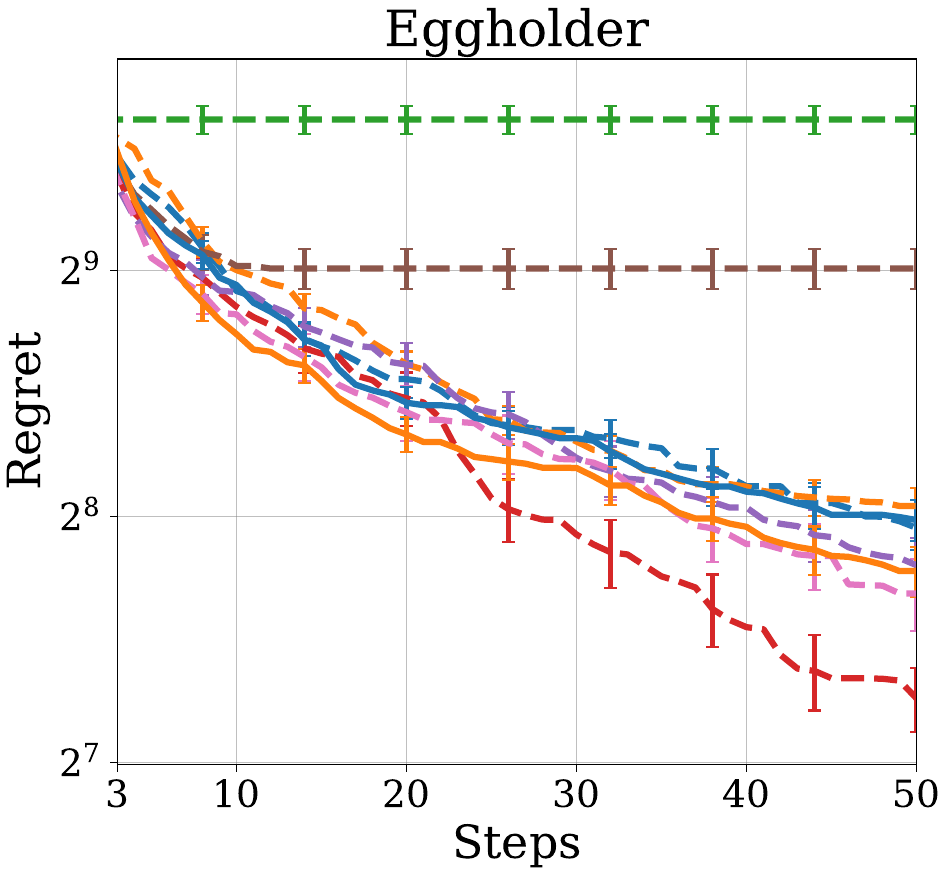}
    \quad
    \includegraphics[width=.75\columnwidth]{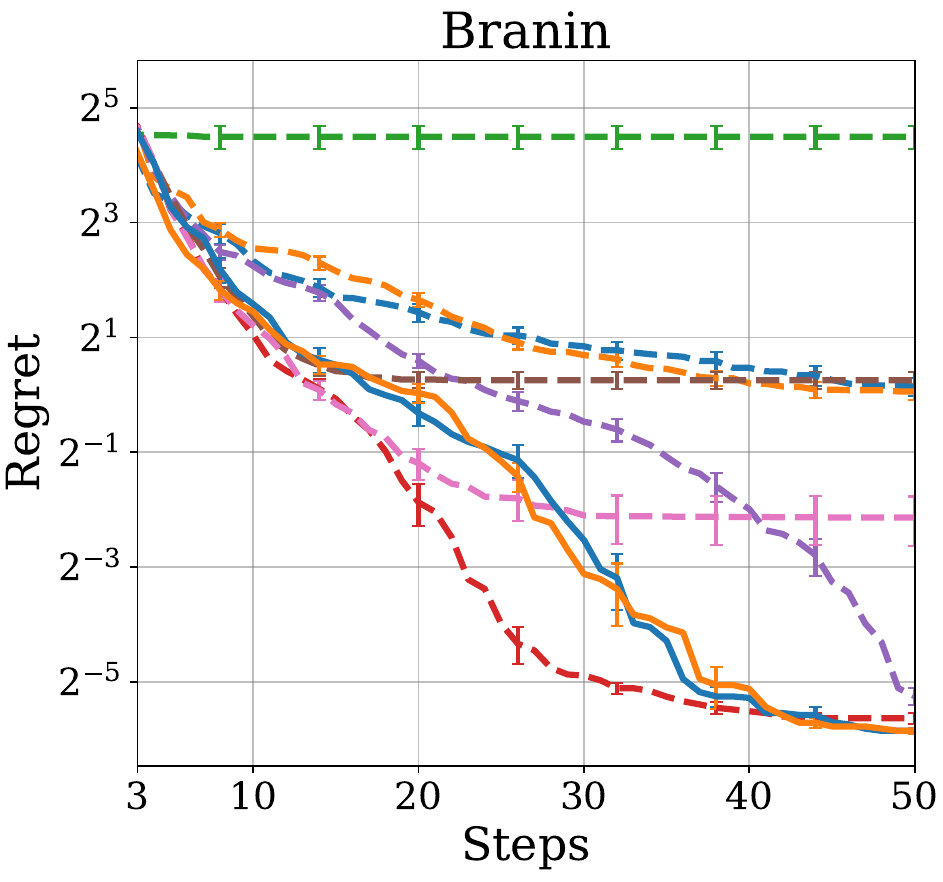}}
    \caption{\label{fg:4bench}Regret measured on 4 standard benchmark functions, averaged over 64 random seeds. The error bars represent one standard deviation of the mean. Two solid lines denote the proposed algorithms and dashed lines denote alternatives. The vertical axis is in log scale.}
\end{figure*}

Figure~\ref{fg:4bench} presents the numerical results on log scale. PI did not work at all, with scores even worse than Random. GP-MI was the second worst one. On the other hand, it is notable that EI works remarkably well especially at later steps. Other methods show more or less similar performance, except that MES tends to easily get stuck in local optima in the Branin function. Overall, the new algorithms---\mbox{{\newbo}-y} and \mbox{{\newbo}-X}---seem to be competitive with others, although the performance ranking appears to be quite sensitive to the kind of the benchmark function.

Table~\ref{tb:4bench} displays averaged cumulative regret 
$\displaystyle \sum_{T=4}^{50}\text{Regret}_T$. It deserves attention that {\newbo}-y outperforms all the other baselines on the Goldstein-Price function. While EI turns out to be best on average across 4 tasks, the three algorithms---MES, {\newbo}-y, and {\newbo}-X---perform almost equally well. At the moment, it is difficult to conclude which of {\newbo}-y and {\newbo}-X is better than the other. 

In summary, the data in Table~\ref{tb:4bench} and Figure~\ref{fg:4bench} indicate that {\newbo} is a robust and effective method of optimizing a black-box function with comparable performance to the state-of-the-art BO methods.

\section{Conclusions and outlook\label{sc:ewtefsd}}

Multi-step decision making benefits from adaptive Bayesian inference \cite{Garnett2023book,Hastie_book2009}. The rising demand for smart sequential measurements of expensive-to-evaluate functions across disciplines, ranging from engineering and medicine to industry, prompts us to develop novel methods for designing a new approach in this field. In this paper, we have put forward a Bayesian optimization scheme dubbed as {\newbo}, which is built upon a measure of statistical dependence between random variables called distance correlation. It is easy to implement, requires no delicate hyperparameter tuning, and performs on par with standard methods such as EI and MES across a variety of black-box maximization tasks. We have also demonstrated the power of {\newbo} in Bayesian experimental design, where the goal is not to determine the optimum of a function but to globally probe the function as accurately as possible with least observations. {\newbo} enables us to execute efficient sequential integral observations with an adaptive integration width in a principled manner. We wish our work finds applications in such fields as computed tomography, remote sensing, and hyperparameter tuning of machine-learning models, where measurements are either costly, time-consuming, or invasive.

One of possible future directions of research is to extend the current work to a batch setup where multiple evaluations are performed in parallel. Another direction is to derive theoretical bounds on properties of the proposed algorithm that complement the empirical analysis of this paper.

\begin{table*}%[tbh]
    \caption{\label{tb:4bench}Cumulative regret $\sum_{T=4}^{50}\text{Regret}_T$ normalized in units of the random policy's mean score. The scores and errors denote the mean and one standard deviation of the mean, over 64 runs with different random seeds. The best scores are in bold, the second scores are underlined, and the third scores are marked with a prime (\scalebox{1.25}{$'$}).}
    \vspace{\baselineskip}
    \centering
    \begin{tabular}{c|cccc}
        & {Goldstein-Price} 
        & {Himmelblau} & {Eggholder} & {Branin} 
        \\\hline \hline 
        \emem{Random} 
        & \underline{1.00 $\pm$ 0.153}
        & 1.00 $\pm$ 0.072
        & 1.00 $\pm$ 0.041
        & 1.00 $\pm$ 0.079
        {\large \strut}\!
        \\
        \emem{VarMax}
        & 2.09 $\pm$ 0.234
        & 1.95 $\pm$ 0.117
        & 1.03 $\pm$ 0.029
        & 1.10 $\pm$ 0.064
        {\large \strut}\!
        \\
        \emem{PI}
        & 71.1 $\pm$ 18.0
        & 9.65 $\pm$ 1.22
        & 2.13 $\pm$ 0.085
        & 7.63 $\pm$ 1.07
        {\large \strut}\!
        \\
        \emem{EI}
        & 1.37 $\pm$ 0.288
        & $\bm{0.448 \pm 0.047}$
        & $\bm{0.828 \pm 0.050}$
        & $\bm{0.400 \pm 0.045}$
        {\large \strut}\!
        \\
        \emem{GP-UCB}
        & 1.38 $\pm$ 0.285
        & 1.20 $\pm$ 0.099
        & 0.960 $\pm$ 0.049
        & 0.724 $\pm$ 0.046
        {\large \strut}\!
        \\
        \emem{GP-MI}
        & 1.86 $\pm$ 0.440
        & 1.58 $\pm$ 0.228
        & 1.42 $\pm$ 0.075
        & 0.725 $\pm$ 0.062
        {\large \strut}\!
        \\
        \emem{MES}
        & 1.39 $\pm$ 0.299
        & \underline{0.657 $\pm$ 0.065}
        & 0.893 $\pm$ 0.058\,\scalebox{1.25}{$'$}
        & 0.455 $\pm$ 0.046\,\scalebox{1.25}{$'$}
        {\large \strut}\!
        \\\hline 
        \emem{\newbo-y}
        & \bm{$0.711 \pm 0.121$}
        & 0.903 $\pm$ 0.066
        & 0.971 $\pm$ 0.034
        & 0.507 $\pm$ 0.044
        {\large \strut}\!
        \\
        \emem{\newbo-X}
        & 1.08 $\pm$ 0.160\,\scalebox{1.25}{$'$}
        & 0.873 $\pm$ 0.077\,\scalebox{1.25}{$'$}
        & \underline{0.880 $\pm$ 0.036}
        & \underline{0.418 $\pm$ 0.028}
        {\large \strut}\!
    \end{tabular}
\end{table*}

\appendix
\section*{\centering Appendix\label{sc:523rwe}}

\begin{itemize}
\item 
Throughout this paper, unless stated otherwise, we consistently used the Mat\'{e}rn 5/2 kernel for GP regression, following the advice of \cite{Snoek2012}. We optimized the length scale and noise level of the kernel following the leave-one-out cross-validation in \cite[Section~5.4.2]{RW_GP_book}. The optimization was done by using a Python library {\sf optuna} \cite{Optuna}.
\item
All our numerical calculations of DC were carried out using a Python library {\sf dcor} \cite{dcor2020,RAMOSCARRENO2023101326}. 
\item 
Although we have also carried out experiments for {\newbo} using \emph{distance covariance} \cite{Szekely2007,Szekely2013,Rizzo2016}, the unnormalized version of DC, we omitted results because the empirical performance with distance covariance turned out to be consistently lower than that using DC.  
\item 
The benchmark functions used in Section~\ref{sc:bench} are defined in the following. Our definitions conform to the convention of \url{https://www.sfu.ca/~ssurjano/index.html} and \url{https://en.wikipedia.org/wiki/Test_functions_for_optimization} to which the reader is referred for further details. Although these functions are usually \emph{minimized} by optimization algorithms, we took the negative of them in Section~\ref{sc:bench} and solved \emph{maximization} problems due to a technical reason.
\end{itemize}
\begin{enumerate}
    \item {\bf Goldstein-Price function}
    \begin{align*}
        f(x_1, x_2) & = \left[1+(x_1+x_2+1)^2 U \right] 
    \\
        & \quad \times \left[30 + (2x_1-3x_2)^2 V \right]
    \end{align*}
    where 
    \begin{align*}
        U & = 19-14x_1+3x_1^2-14x_2+6x_1x_2+3x_2^2 \,,
        \\
        V & = 18-32x_1+12x_1^2+48x_2-36x_1x_2+27x_2^2 \,.
    \end{align*}
    The domain is $-2\leq x_1 \leq 2$ and $-2\leq x_2 \leq 2$. The minimum value is $3$.
    %%%%%%%%%%%
    %%%%%%%%%%%
    \item {\bf Himmelblau function}
    \[
        f(x_1, x_2) = (x_1^2+x_2-11)^2 + (x_1+x_2^2-7)^2.
    \]
    The domain is $-6\leq x_1\leq 6$ and $-6\leq x_2\leq 6$. The minimum value is $0$.
    %%%%%%%%%%%
    %%%%%%%%%%%
    \item {\bf Eggholder function}
    \[
        \begin{array}{rl}
        f(x_1, x_2) = \!\!\!\! & - (x_2+47) \sin 
        \left(\sqrt{\left|x_2+\frac{x_1}{2}+47\right|}\right) 
        \\
        & - x_1 \sin \left(\sqrt{|x_1-(x_2+47)|}\right). 
        \end{array}
    \]
        The domain is $-512\leq x_1\leq 512$ and $-512\leq x_2\leq 512$. The minimum value is $-959.640663$.
    %%%%%%%%%%%
    %%%%%%%%%%%
    \item {\bf Branin (or Branin-Hoo) function}
    \[
        f(x_1, x_2) = a(x_2-bx_1^2+cx_1-r)^2 + s(1-t)\cos x_1 + s
    \]
    where $a=1$, $b=5.1/(4\pi^2)$, $c=5/\pi$, $r=6$, $s=10$ and $t=1/(8\pi)$.
    The domain is $-5 \leq x_1 \leq 10$ and $0 \leq x_2 \leq 15$. The minimum value is $0.39788736$.
\end{enumerate}

%\bibliographystyle{model1-num-names}
%\bibliography{refs_bo}
\bibliography{draft_bo_for_arXiv.bbl}

\begin{thebibliography}{70}
\expandafter\ifx\csname natexlab\endcsname\relax\def\natexlab#1{#1}\fi
\providecommand{\bibinfo}[2]{#2}
\ifx\xfnm\relax \def\xfnm[#1]{\unskip,\space#1}\fi
%Type = Book
\bibitem[{Bertsekas(2017)}]{Bertsekas2017}
\bibinfo{author}{D.~P. Bertsekas}, \bibinfo{title}{{{Dynamic Programming and
  Optimal Control}}}, \bibinfo{publisher}{Athena Scientific},
  \bibinfo{edition}{4th} edition, \bibinfo{year}{2017}.
%Type = Book
\bibitem[{Sutton and Barto(2018)}]{SuttonBarto}
\bibinfo{author}{R.~S. Sutton}, \bibinfo{author}{A.~G. Barto},
  \bibinfo{title}{{{Reinforcement Learning: An Introduction}}},
  \bibinfo{publisher}{MIT Press}, \bibinfo{year}{2018}.
%Type = Book
\bibitem[{Garnett(2023)}]{Garnett2023book}
\bibinfo{author}{R.~Garnett}, \bibinfo{title}{{Bayesian Optimization}},
  \bibinfo{publisher}{Cambridge University Press}, \bibinfo{year}{2023}.
%Type = Article
\bibitem[{Brochu et~al.(2010)Brochu, Cora, and de~Freitas}]{Brochu2010tutorial}
\bibinfo{author}{E.~Brochu}, \bibinfo{author}{V.~M. Cora},
  \bibinfo{author}{N.~de~Freitas},
\newblock \bibinfo{title}{{{A Tutorial on Bayesian Optimization of Expensive
  Cost Functions, with Application to Active User Modeling and Hierarchical
  Reinforcement Learning}}}  (\bibinfo{year}{2010}).
  \bibinfo{note}{ArXiv:1012.2599}.
%Type = Article
\bibitem[{{Shahriari} et~al.(2016){Shahriari}, {Swersky}, {Wang}, {Adams}, and
  {de Freitas}}]{Shahriari2016}
\bibinfo{author}{B.~{Shahriari}}, \bibinfo{author}{K.~{Swersky}},
  \bibinfo{author}{Z.~{Wang}}, \bibinfo{author}{R.~P. {Adams}},
  \bibinfo{author}{N.~{de Freitas}},
\newblock \bibinfo{title}{{{Taking the Human Out of the Loop: A Review of
  Bayesian Optimization}}},
\newblock \bibinfo{journal}{Proceedings of the IEEE} \bibinfo{volume}{104}
  (\bibinfo{year}{2016}) \bibinfo{pages}{148--175}.
%Type = Article
\bibitem[{Frazier(2018)}]{Frazier2018tutorial}
\bibinfo{author}{P.~I. Frazier},
\newblock \bibinfo{title}{{{A Tutorial on Bayesian Optimization}}}
  (\bibinfo{year}{2018}). \bibinfo{note}{ArXiv:1807.02811}.
%Type = Misc
\bibitem[{Li(2017)}]{Li2017}
\bibinfo{author}{Y.~Li}, \bibinfo{title}{{{Deep Reinforcement Learning: An
  Overview}}}, \bibinfo{year}{2017}. \bibinfo{note}{ArXiv:1701.07274}.
%Type = Article
\bibitem[{François-Lavet et~al.(2018)François-Lavet, Henderson, Islam,
  Bellemare, and Pineau}]{Franois_Lavet_2018}
\bibinfo{author}{V.~François-Lavet}, \bibinfo{author}{P.~Henderson},
  \bibinfo{author}{R.~Islam}, \bibinfo{author}{M.~G. Bellemare},
  \bibinfo{author}{J.~Pineau},
\newblock \bibinfo{title}{{{An Introduction to Deep Reinforcement Learning}}},
\newblock \bibinfo{journal}{Foundations and Trends in Machine Learning}
  \bibinfo{volume}{11} (\bibinfo{year}{2018}) \bibinfo{pages}{219--354}.
%Type = Article
\bibitem[{Kushner(1964)}]{Kushner1964}
\bibinfo{author}{H.~J. Kushner},
\newblock \bibinfo{title}{{A New Method of Locating the Maximum Point of an
  Arbitrary Multipeak Curve in the Presence of Noise}},
\newblock \bibinfo{journal}{Journal of Basic Engineering} \bibinfo{volume}{86}
  (\bibinfo{year}{1964}) \bibinfo{pages}{97--106}.
%Type = Article
\bibitem[{Jones et~al.(1998)Jones, Schonlau, and Welch}]{Jones1998}
\bibinfo{author}{D.~R. Jones}, \bibinfo{author}{M.~Schonlau},
  \bibinfo{author}{W.~J. Welch},
\newblock \bibinfo{title}{{{Efficient global optimization of expensive
  black-box functions}}},
\newblock \bibinfo{journal}{Journal of Global Optimization}
  \bibinfo{volume}{13} (\bibinfo{year}{1998}) \bibinfo{pages}{455--492}.
%Type = Article
\bibitem[{{Srinivas} et~al.(2012){Srinivas}, {Krause}, {Kakade}, and
  {Seeger}}]{Srinivas2012}
\bibinfo{author}{N.~{Srinivas}}, \bibinfo{author}{A.~{Krause}},
  \bibinfo{author}{S.~M. {Kakade}}, \bibinfo{author}{M.~W. {Seeger}},
\newblock \bibinfo{title}{{{Information-Theoretic Regret Bounds for Gaussian
  Process Optimization in the Bandit Setting}}},
\newblock \bibinfo{journal}{IEEE Transactions on Information Theory}
  \bibinfo{volume}{58} (\bibinfo{year}{2012}) \bibinfo{pages}{3250--3265}.
%Type = Inproceedings
\bibitem[{Contal et~al.(2014)Contal, Perchet, and Vayatis}]{Contal2014}
\bibinfo{author}{E.~Contal}, \bibinfo{author}{V.~Perchet},
  \bibinfo{author}{N.~Vayatis},
\newblock \bibinfo{title}{{{Gaussian Process Optimization with Mutual
  Information}}},
\newblock in: \bibinfo{booktitle}{Proceedings of the 31st International
  Conference on Machine Learning}, volume~\bibinfo{volume}{32} of
  \textit{\bibinfo{series}{PMLR}}, \bibinfo{publisher}{PMLR},
  \bibinfo{year}{2014}, pp. \bibinfo{pages}{253--261}.
%Type = Inproceedings
\bibitem[{Wang and Jegelka(2017)}]{Wang2017}
\bibinfo{author}{Z.~Wang}, \bibinfo{author}{S.~Jegelka},
\newblock \bibinfo{title}{{{Max-Value Entropy Search for Efficient Bayesian
  Optimization}}},
\newblock in: \bibinfo{booktitle}{Proceedings of the 34th International
  Conference on Machine Learning}, volume~\bibinfo{volume}{70}, pp.
  \bibinfo{pages}{3627--3635}.
%Type = Inproceedings
\bibitem[{Kanazawa(2022)}]{kanazawa2022ijcnn}
\bibinfo{author}{T.~Kanazawa},
\newblock \bibinfo{title}{{{One-Parameter Family of New Acquisition Functions
  for Efficient Global Optimization}}},
\newblock in: \bibinfo{booktitle}{2022 International Joint Conference on Neural
  Networks (IJCNN)}, pp. \bibinfo{pages}{1--8}.
%Type = Article
\bibitem[{Székely et~al.(2007)Székely, Rizzo, and Bakirov}]{Szekely2007}
\bibinfo{author}{G.~J. Székely}, \bibinfo{author}{M.~L. Rizzo},
  \bibinfo{author}{N.~K. Bakirov},
\newblock \bibinfo{title}{{{Measuring and Testing Dependence by Correlation of
  Distances}}},
\newblock \bibinfo{journal}{Annals of Statistics} \bibinfo{volume}{35}
  (\bibinfo{year}{2007}) \bibinfo{pages}{2769--2794}.
%Type = Article
\bibitem[{Székely and Rizzo(2013)}]{Szekely2013}
\bibinfo{author}{G.~J. Székely}, \bibinfo{author}{M.~L. Rizzo},
\newblock \bibinfo{title}{{{Energy statistics: A class of statistics based on
  distances}}},
\newblock \bibinfo{journal}{Journal of Statistical Planning and Inference}
  \bibinfo{volume}{143} (\bibinfo{year}{2013}) \bibinfo{pages}{1249--1272}.
%Type = Article
\bibitem[{Rizzo and Székely(2016)}]{Rizzo2016}
\bibinfo{author}{M.~L. Rizzo}, \bibinfo{author}{G.~J. Székely},
\newblock \bibinfo{title}{{Energy distance}},
\newblock \bibinfo{journal}{WIREs Comput Stat} \bibinfo{volume}{8}
  (\bibinfo{year}{2016}) \bibinfo{pages}{27--38}.
%Type = Article
\bibitem[{Maimaitijiang et~al.(2020)Maimaitijiang, Sagan, Sidike, Daloye,
  Erkbol, and Fritschi}]{rs12091357}
\bibinfo{author}{M.~Maimaitijiang}, \bibinfo{author}{V.~Sagan},
  \bibinfo{author}{P.~Sidike}, \bibinfo{author}{A.~M. Daloye},
  \bibinfo{author}{H.~Erkbol}, \bibinfo{author}{F.~B. Fritschi},
\newblock \bibinfo{title}{{{Crop Monitoring Using Satellite/UAV Data Fusion and
  Machine Learning}}},
\newblock \bibinfo{journal}{Remote Sensing} \bibinfo{volume}{12}
  (\bibinfo{year}{2020}) \bibinfo{pages}{1357}.
%Type = Article
\bibitem[{Burke et~al.(2021)Burke, Driscoll, Lobell, and Ermon}]{Burke_2021}
\bibinfo{author}{M.~Burke}, \bibinfo{author}{A.~Driscoll},
  \bibinfo{author}{D.~B. Lobell}, \bibinfo{author}{S.~Ermon},
\newblock \bibinfo{title}{{Using satellite imagery to understand and promote
  sustainable development}},
\newblock \bibinfo{journal}{Science} \bibinfo{volume}{371}
  (\bibinfo{year}{2021}) \bibinfo{pages}{eabe8628}.
%Type = Article
\bibitem[{Benos et~al.(2021)Benos, Tagarakis, Dolias, Berruto, Kateris, and
  Bochtis}]{s21113758}
\bibinfo{author}{L.~Benos}, \bibinfo{author}{A.~C. Tagarakis},
  \bibinfo{author}{G.~Dolias}, \bibinfo{author}{R.~Berruto},
  \bibinfo{author}{D.~Kateris}, \bibinfo{author}{D.~Bochtis},
\newblock \bibinfo{title}{{{Machine Learning in Agriculture: A Comprehensive
  Updated Review}}},
\newblock \bibinfo{journal}{Sensors} \bibinfo{volume}{21}
  (\bibinfo{year}{2021}) \bibinfo{pages}{3758}.
%Type = Inproceedings
\bibitem[{Uzkent and Ermon(2020)}]{Uzkent2020}
\bibinfo{author}{B.~Uzkent}, \bibinfo{author}{S.~Ermon},
\newblock \bibinfo{title}{{{Learning When and Where to Zoom with Deep
  Reinforcement Learning}}},
\newblock in: \bibinfo{booktitle}{Proceedings of the IEEE/CVF Conference on
  Computer Vision and Pattern Recognition (CVPR)}, pp.
  \bibinfo{pages}{12345--12354}. \bibinfo{note}{ArXiv:2003.00425}.
%Type = Article
\bibitem[{Ayush et~al.(2020)Ayush, Uzkent, Burke, Lobell, and
  Ermon}]{Ayush2020}
\bibinfo{author}{K.~Ayush}, \bibinfo{author}{B.~Uzkent},
  \bibinfo{author}{M.~Burke}, \bibinfo{author}{D.~Lobell},
  \bibinfo{author}{S.~Ermon},
\newblock \bibinfo{title}{{{Efficient Poverty Mapping using Deep Reinforcement
  Learning}}}  (\bibinfo{year}{2020}). \bibinfo{note}{ArXiv:2006.04224}.
%Type = Book
\bibitem[{Hastie et~al.(2009)Hastie, Tibshirani, and
  Friedman}]{Hastie_book2009}
\bibinfo{author}{T.~Hastie}, \bibinfo{author}{R.~Tibshirani},
  \bibinfo{author}{J.~Friedman}, \bibinfo{title}{{{The Elements of Statistical
  Learning}}}, \bibinfo{publisher}{Springer New York}, \bibinfo{address}{NY},
  \bibinfo{year}{2009}.
%Type = Article
\bibitem[{Reshef et~al.(2011)Reshef, Reshef, Finucane, Grossman, McVean,
  Turnbaugh, Lander, Mitzenmacher, and Sabeti}]{Reshef2011}
\bibinfo{author}{D.~N. Reshef}, \bibinfo{author}{Y.~A. Reshef},
  \bibinfo{author}{H.~K. Finucane}, \bibinfo{author}{S.~R. Grossman},
  \bibinfo{author}{G.~McVean}, \bibinfo{author}{P.~J. Turnbaugh},
  \bibinfo{author}{E.~S. Lander}, \bibinfo{author}{M.~Mitzenmacher},
  \bibinfo{author}{P.~C. Sabeti},
\newblock \bibinfo{title}{{Detecting Novel Associations in Large Data Sets}},
\newblock \bibinfo{journal}{Science} \bibinfo{volume}{334}
  (\bibinfo{year}{2011}) \bibinfo{pages}{1518--1524}.
%Type = Inproceedings
\bibitem[{Gretton et~al.(2005)Gretton, Bousquet, Smola, and
  Sch{\"o}lkopf}]{HSIC2005}
\bibinfo{author}{A.~Gretton}, \bibinfo{author}{O.~Bousquet},
  \bibinfo{author}{A.~Smola}, \bibinfo{author}{B.~Sch{\"o}lkopf},
\newblock \bibinfo{title}{{Measuring Statistical Dependence with
  Hilbert-Schmidt Norms}},
\newblock in: \bibinfo{editor}{S.~Jain}, \bibinfo{editor}{H.~U. Simon},
  \bibinfo{editor}{E.~Tomita} (Eds.), \bibinfo{booktitle}{{Algorithmic Learning
  Theory}}, \bibinfo{publisher}{Springer Berlin Heidelberg},
  \bibinfo{address}{Berlin, Heidelberg}, \bibinfo{year}{2005}, pp.
  \bibinfo{pages}{63--77}.
%Type = Inproceedings
\bibitem[{Gretton et~al.(2007{\natexlab{a}})Gretton, Fukumizu, Teo, Song,
  Sch\"{o}lkopf, and Smola}]{HSIC2007}
\bibinfo{author}{A.~Gretton}, \bibinfo{author}{K.~Fukumizu},
  \bibinfo{author}{C.~Teo}, \bibinfo{author}{L.~Song},
  \bibinfo{author}{B.~Sch\"{o}lkopf}, \bibinfo{author}{A.~Smola},
\newblock \bibinfo{title}{{{A Kernel Statistical Test of Independence}}},
\newblock in: \bibinfo{editor}{J.~Platt}, \bibinfo{editor}{D.~Koller},
  \bibinfo{editor}{Y.~Singer}, \bibinfo{editor}{S.~Roweis} (Eds.),
  \bibinfo{booktitle}{Advances in Neural Information Processing Systems},
  volume~\bibinfo{volume}{20}, \bibinfo{publisher}{Curran Associates, Inc.},
  \bibinfo{year}{2007}{\natexlab{a}}, pp. \bibinfo{pages}{585--592}.
%Type = Inproceedings
\bibitem[{Gretton et~al.(2007{\natexlab{b}})Gretton, Borgwardt, Rasch,
  Sch\"{o}lkopf, and Smola}]{Gretton_NIPS2006}
\bibinfo{author}{A.~Gretton}, \bibinfo{author}{K.~Borgwardt},
  \bibinfo{author}{M.~Rasch}, \bibinfo{author}{B.~Sch\"{o}lkopf},
  \bibinfo{author}{A.~Smola},
\newblock \bibinfo{title}{{A Kernel Method for the Two-Sample-Problem}},
\newblock in: \bibinfo{editor}{B.~Sch\"{o}lkopf}, \bibinfo{editor}{J.~Platt},
  \bibinfo{editor}{T.~Hoffman} (Eds.), \bibinfo{booktitle}{Advances in Neural
  Information Processing Systems}, volume~\bibinfo{volume}{19},
  \bibinfo{publisher}{MIT Press}, \bibinfo{year}{2007}{\natexlab{b}}, pp.
  \bibinfo{pages}{513--520}.
%Type = Article
\bibitem[{Sejdinovic et~al.(2013)Sejdinovic, Sriperumbudur, Gretton, and
  Fukumizu}]{Sejdinovic2013}
\bibinfo{author}{D.~Sejdinovic}, \bibinfo{author}{B.~Sriperumbudur},
  \bibinfo{author}{A.~Gretton}, \bibinfo{author}{K.~Fukumizu},
\newblock \bibinfo{title}{{Equivalence of distance-based and RKHS-based
  statistics in hypothesis testing}},
\newblock \bibinfo{journal}{The Annals of Statistics} \bibinfo{volume}{41}
  (\bibinfo{year}{2013}) \bibinfo{pages}{2263--2291}.
%Type = Inproceedings
\bibitem[{Zhen et~al.(2022)Zhen, Meng, Chakraborty, and Singh}]{Zhen2022}
\bibinfo{author}{X.~Zhen}, \bibinfo{author}{Z.~Meng},
  \bibinfo{author}{R.~Chakraborty}, \bibinfo{author}{V.~Singh},
\newblock \bibinfo{title}{{{On the Versatile Uses of Partial Distance
  Correlation in Deep Learning}}},
\newblock in: \bibinfo{editor}{S.~Avidan}, \bibinfo{editor}{G.~Brostow},
  \bibinfo{editor}{M.~Ciss{\'e}}, \bibinfo{editor}{G.~M. Farinella},
  \bibinfo{editor}{T.~Hassner} (Eds.), \bibinfo{booktitle}{Computer Vision --
  ECCV 2022}, \bibinfo{publisher}{Springer Nature Switzerland},
  \bibinfo{address}{Cham}, \bibinfo{year}{2022}, pp. \bibinfo{pages}{327--346}.
%Type = Book
\bibitem[{Rasmussen and Williams(2006)}]{RW_GP_book}
\bibinfo{author}{C.~E. Rasmussen}, \bibinfo{author}{C.~K.~I. Williams},
  \bibinfo{title}{{{Gaussian Processes for Machine Learning}}},
  \bibinfo{publisher}{MIT Press}, \bibinfo{year}{2006}.
%Type = Inproceedings
\bibitem[{Solak et~al.(2002)Solak, Murray-Smith, Leithead, Leith, and
  Rasmussen}]{Solak2002nips}
\bibinfo{author}{E.~Solak}, \bibinfo{author}{R.~Murray-Smith},
  \bibinfo{author}{W.~E. Leithead}, \bibinfo{author}{D.~J. Leith},
  \bibinfo{author}{C.~E. Rasmussen},
\newblock \bibinfo{title}{{{Derivative Observations in Gaussian Process Models
  of Dynamic Systems}}},
\newblock in: \bibinfo{booktitle}{Proceedings of the 15th International
  Conference on Neural Information Processing Systems}, \bibinfo{publisher}{MIT
  Press}, \bibinfo{year}{2002}, pp. \bibinfo{pages}{1057--1064}.
%Type = Inproceedings
\bibitem[{Osborne et~al.(2009)Osborne, Garnett, and Roberts}]{Osborne2009}
\bibinfo{author}{M.~A. Osborne}, \bibinfo{author}{R.~Garnett},
  \bibinfo{author}{S.~J. Roberts},
\newblock \bibinfo{title}{Gaussian processes for global optimization},
\newblock in: \bibinfo{booktitle}{3rd International Conference on Learning and
  Intelligent Optimization (LION3)}.
%Type = Article
\bibitem[{Ahmed et~al.(2016)Ahmed, Shahriari, and Schmidt}]{Ahmed2016}
\bibinfo{author}{M.~Ahmed}, \bibinfo{author}{B.~Shahriari},
  \bibinfo{author}{M.~Schmidt},
\newblock \bibinfo{title}{{{Do we need ``harmless'' Bayesian optimization and
  ``first-order'' Bayesian optimization?}}},
\newblock \bibinfo{journal}{BayesOpt}  (\bibinfo{year}{2016}).
%Type = Inproceedings
\bibitem[{Wu et~al.(2017{\natexlab{a}})Wu, Poloczek, Wilson, and
  Frazier}]{Wu2017nips}
\bibinfo{author}{J.~Wu}, \bibinfo{author}{M.~Poloczek}, \bibinfo{author}{A.~G.
  Wilson}, \bibinfo{author}{P.~Frazier},
\newblock \bibinfo{title}{{{Bayesian Optimization with Gradients}}},
\newblock in: \bibinfo{booktitle}{Advances in Neural Information Processing
  Systems}, volume~\bibinfo{volume}{30}, pp. \bibinfo{pages}{5267--5278}.
%Type = Misc
\bibitem[{Wu et~al.(2017{\natexlab{b}})Wu, Aoi, and Pillow}]{Wu2017arXiv}
\bibinfo{author}{A.~Wu}, \bibinfo{author}{M.~C. Aoi}, \bibinfo{author}{J.~W.
  Pillow}, \bibinfo{title}{{{Exploiting gradients and Hessians in Bayesian
  optimization and Bayesian quadrature}}}, \bibinfo{year}{2017}{\natexlab{b}}.
  \bibinfo{note}{ArXiv:1704.00060}.
%Type = Incollection
\bibitem[{Eriksson et~al.(2018)Eriksson, Dong, Lee, Bindel, and
  Wilson}]{Eriksson2018nips}
\bibinfo{author}{D.~Eriksson}, \bibinfo{author}{K.~Dong},
  \bibinfo{author}{E.~Lee}, \bibinfo{author}{D.~Bindel}, \bibinfo{author}{A.~G.
  Wilson},
\newblock \bibinfo{title}{{{Scaling Gaussian Process Regression with
  Derivatives}}},
\newblock in: \bibinfo{booktitle}{Advances in Neural Information Processing
  Systems 31}, \bibinfo{year}{2018}, pp. \bibinfo{pages}{6867--6877}.
%Type = Misc
\bibitem[{Smith et~al.(2018)Smith, Alvarez, and Lawrence}]{Smith2018}
\bibinfo{author}{M.~T. Smith}, \bibinfo{author}{M.~A. Alvarez},
  \bibinfo{author}{N.~D. Lawrence}, \bibinfo{title}{{{Gaussian Process
  Regression for Binned Data}}}, \bibinfo{year}{2018}.
  \bibinfo{note}{ArXiv:1809.02010}.
%Type = Inproceedings
\bibitem[{Adelsberg and Schwantes(2018)}]{Adelsberg2018}
\bibinfo{author}{M.~Adelsberg}, \bibinfo{author}{C.~Schwantes},
\newblock \bibinfo{title}{{{Binned Kernels for Anomaly Detection in
  Multi-timescale Data using Gaussian Processes}}},
\newblock in: \bibinfo{booktitle}{Proceedings of the KDD 2017: Workshop on
  Anomaly Detection in Finance}, volume~\bibinfo{volume}{71} of
  \textit{\bibinfo{series}{PMLR}}, pp. \bibinfo{pages}{102--113}.
%Type = Inproceedings
\bibitem[{Law et~al.(2018)Law, Sejdinovic, Cameron, Lucas, Flaxman, Battle, and
  Fukumizu}]{Law2018nips}
\bibinfo{author}{H.~C.~L. Law}, \bibinfo{author}{D.~Sejdinovic},
  \bibinfo{author}{E.~Cameron}, \bibinfo{author}{T.~C.~D. Lucas},
  \bibinfo{author}{S.~Flaxman}, \bibinfo{author}{K.~Battle},
  \bibinfo{author}{K.~Fukumizu},
\newblock \bibinfo{title}{{{Variational learning on aggregate outputs with
  Gaussian processes}}},
\newblock in: \bibinfo{booktitle}{Proceedings of the 32nd International
  Conference on Neural Information Processing Systems}, pp.
  \bibinfo{pages}{6084--6094}.
%Type = Misc
\bibitem[{Hendriks et~al.(2018)Hendriks, Jidling, Wills, and
  Schön}]{Hendriks2018}
\bibinfo{author}{J.~N. Hendriks}, \bibinfo{author}{C.~Jidling},
  \bibinfo{author}{A.~Wills}, \bibinfo{author}{T.~B. Schön},
  \bibinfo{title}{{{Evaluating the squared-exponential covariance function in
  Gaussian processes with integral observations}}}, \bibinfo{year}{2018}.
  \bibinfo{note}{ArXiv:1812.07319}.
%Type = Article
\bibitem[{Purisha et~al.(2019)Purisha, Jidling, Wahlström, Schön, and
  Särkkä}]{Purisha2019}
\bibinfo{author}{Z.~Purisha}, \bibinfo{author}{C.~Jidling},
  \bibinfo{author}{N.~Wahlström}, \bibinfo{author}{T.~B. Schön},
  \bibinfo{author}{S.~Särkkä},
\newblock \bibinfo{title}{Probabilistic approach to limited-data computed
  tomography reconstruction},
\newblock \bibinfo{journal}{Inverse Problems} \bibinfo{volume}{35}
  (\bibinfo{year}{2019}) \bibinfo{pages}{105004}.
%Type = Misc
\bibitem[{Jidling et~al.(2019)Jidling, Hendriks, Schön, and
  Wills}]{Jidling2019}
\bibinfo{author}{C.~Jidling}, \bibinfo{author}{J.~Hendriks},
  \bibinfo{author}{T.~B. Schön}, \bibinfo{author}{A.~Wills},
  \bibinfo{title}{Deep kernel learning for integral measurements},
  \bibinfo{year}{2019}. \bibinfo{note}{ArXiv:1909.01844}.
%Type = Inproceedings
\bibitem[{Hamelijnck et~al.(2019)Hamelijnck, Damoulas, Wang, and
  Girolami}]{Hamelijnck2019nips}
\bibinfo{author}{O.~Hamelijnck}, \bibinfo{author}{T.~Damoulas},
  \bibinfo{author}{K.~Wang}, \bibinfo{author}{M.~Girolami},
\newblock \bibinfo{title}{{{Multi-resolution Multi-task Gaussian Processes}}},
\newblock in: \bibinfo{booktitle}{Advances in Neural Information Processing
  Systems}, volume~\bibinfo{volume}{32}, pp. \bibinfo{pages}{14025--14035}.
%Type = Inproceedings
\bibitem[{Tanaka et~al.(2019)Tanaka, Tanaka, Iwata, Kurashima, Okawa, Akagi,
  and Toda}]{Tanaka2019nips}
\bibinfo{author}{Y.~Tanaka}, \bibinfo{author}{T.~Tanaka},
  \bibinfo{author}{T.~Iwata}, \bibinfo{author}{T.~Kurashima},
  \bibinfo{author}{M.~Okawa}, \bibinfo{author}{Y.~Akagi},
  \bibinfo{author}{H.~Toda},
\newblock \bibinfo{title}{{{Spatially Aggregated Gaussian Processes with
  Multivariate Areal Outputs}}},
\newblock in: \bibinfo{booktitle}{Advances in Neural Information Processing
  Systems}, volume~\bibinfo{volume}{32}, pp. \bibinfo{pages}{3005--3015}.
%Type = Inproceedings
\bibitem[{Yousefi et~al.(2019)Yousefi, Smith, and
  \'{A}lvarez}]{Yousefi2019nips}
\bibinfo{author}{F.~Yousefi}, \bibinfo{author}{M.~T. Smith},
  \bibinfo{author}{M.~\'{A}lvarez},
\newblock \bibinfo{title}{{{Multi-task Learning for Aggregated Data using
  Gaussian Processes}}},
\newblock in: \bibinfo{booktitle}{Advances in Neural Information Processing
  Systems}, volume~\bibinfo{volume}{32}, pp. \bibinfo{pages}{15076--15086}.
%Type = Inproceedings
\bibitem[{{Tanskanen} et~al.(2020){Tanskanen}, {Longi}, and
  {Klami}}]{Tanskanen2020}
\bibinfo{author}{V.~{Tanskanen}}, \bibinfo{author}{K.~{Longi}},
  \bibinfo{author}{A.~{Klami}},
\newblock \bibinfo{title}{{{Non-Linearities in Gaussian Processes with Integral
  Observations}}},
\newblock in: \bibinfo{booktitle}{2020 IEEE 30th International Workshop on
  Machine Learning for Signal Processing (MLSP)}, pp. \bibinfo{pages}{1--6}.
%Type = Inproceedings
\bibitem[{Longi et~al.(2020)Longi, Rajani, Sillanp\"{a}\"{a}, M\"{a}kinen,
  Rauhala, Salmi, Haeggstr\"{o}m, and Klami}]{Longi2020}
\bibinfo{author}{K.~Longi}, \bibinfo{author}{C.~Rajani},
  \bibinfo{author}{T.~Sillanp\"{a}\"{a}}, \bibinfo{author}{J.~M\"{a}kinen},
  \bibinfo{author}{T.~Rauhala}, \bibinfo{author}{A.~Salmi},
  \bibinfo{author}{E.~Haeggstr\"{o}m}, \bibinfo{author}{A.~Klami},
\newblock \bibinfo{title}{{{Sensor Placement for Spatial Gaussian Processes
  with Integral Observations}}},
\newblock in: \bibinfo{booktitle}{Proceedings of the 36th Conference on
  Uncertainty in Artificial Intelligence (UAI)}, volume \bibinfo{volume}{124}
  of \textit{\bibinfo{series}{PMLR}}, pp. \bibinfo{pages}{1009--1018}.
%Type = Article
\bibitem[{Villemonteix et~al.(2009)Villemonteix, Vazquez, and
  Walter}]{Villemonteix2009}
\bibinfo{author}{J.~Villemonteix}, \bibinfo{author}{E.~Vazquez},
  \bibinfo{author}{E.~Walter},
\newblock \bibinfo{title}{{{An informational approach to the global
  optimization of expensive-to-evaluate functions}}},
\newblock \bibinfo{journal}{Journal of Global Optimization}
  \bibinfo{volume}{44} (\bibinfo{year}{2009}) \bibinfo{pages}{509}.
%Type = Article
\bibitem[{Hennig and Schuler(2012)}]{Hennig2012}
\bibinfo{author}{P.~Hennig}, \bibinfo{author}{C.~J. Schuler},
\newblock \bibinfo{title}{{{Entropy Search for Information-Efficient Global
  Optimization}}},
\newblock \bibinfo{journal}{Journal of Machine Learning Research}
  \bibinfo{volume}{13} (\bibinfo{year}{2012}) \bibinfo{pages}{1809--1837}.
%Type = Inproceedings
\bibitem[{Hern\'{a}ndez-Lobato et~al.(2014)Hern\'{a}ndez-Lobato, Hoffman, and
  Ghahramani}]{HernandezLobato2014}
\bibinfo{author}{J.~M. Hern\'{a}ndez-Lobato}, \bibinfo{author}{M.~W. Hoffman},
  \bibinfo{author}{Z.~Ghahramani},
\newblock \bibinfo{title}{{{Predictive Entropy Search for Efficient Global
  Optimization of Black-box Functions}}},
\newblock in: \bibinfo{booktitle}{Advances in Neural Information Processing
  Systems}, volume~\bibinfo{volume}{27}, pp. \bibinfo{pages}{918--926}.
%Type = Article
\bibitem[{Hoffman and Ghahramani(2015)}]{Hoffman2015}
\bibinfo{author}{M.~W. Hoffman}, \bibinfo{author}{Z.~Ghahramani},
\newblock \bibinfo{title}{{{Output-space predictive entropy search for flexible
  global optimization}}},
\newblock \bibinfo{journal}{NIPS workshop on Bayesian Optimization}
  (\bibinfo{year}{2015}).
%Type = Inproceedings
\bibitem[{Takeno et~al.(2020)Takeno, Fukuoka, Tsukada, Koyama, Shiga, Takeuchi,
  and Karasuyama}]{Takeno2020}
\bibinfo{author}{S.~Takeno}, \bibinfo{author}{H.~Fukuoka},
  \bibinfo{author}{Y.~Tsukada}, \bibinfo{author}{T.~Koyama},
  \bibinfo{author}{M.~Shiga}, \bibinfo{author}{I.~Takeuchi},
  \bibinfo{author}{M.~Karasuyama},
\newblock \bibinfo{title}{{{Multi-fidelity Bayesian Optimization with Max-value
  Entropy Search and its Parallelization}}},
\newblock in: \bibinfo{booktitle}{Proceedings of the 37th International
  Conference on Machine Learning}, volume \bibinfo{volume}{119} of
  \textit{\bibinfo{series}{PMLR}}, pp. \bibinfo{pages}{9334--9345}.
%Type = Inproceedings
\bibitem[{Moss et~al.(2020)Moss, Leslie, and Rayson}]{Moss2020}
\bibinfo{author}{H.~B. Moss}, \bibinfo{author}{D.~S. Leslie},
  \bibinfo{author}{P.~Rayson},
\newblock \bibinfo{title}{{{MUMBO: MUlti-task Max-value Bayesian
  Optimization}}},
\newblock in: \bibinfo{booktitle}{Proceedings of Machine Learning and Knowledge
  Discovery in Databases: European Conference, ECML PKDD 2020, Part III},
  \bibinfo{address}{Ghent, Belgium}, pp. \bibinfo{pages}{447--462}.
  \bibinfo{note}{ArXiv:2006.12093}.
%Type = Inproceedings
\bibitem[{Belakaria et~al.(2019)Belakaria, Deshwal, and
  Doppa}]{NEURIPS2019_82edc5c9}
\bibinfo{author}{S.~Belakaria}, \bibinfo{author}{A.~Deshwal},
  \bibinfo{author}{J.~R. Doppa},
\newblock \bibinfo{title}{{{Max-value Entropy Search for Multi-Objective
  Bayesian Optimization}}},
\newblock in: \bibinfo{editor}{H.~Wallach}, \bibinfo{editor}{H.~Larochelle},
  \bibinfo{editor}{A.~Beygelzimer}, \bibinfo{editor}{F.~d\textquotesingle
  Alch\'{e}-Buc}, \bibinfo{editor}{E.~Fox}, \bibinfo{editor}{R.~Garnett}
  (Eds.), \bibinfo{booktitle}{Advances in Neural Information Processing
  Systems}, volume~\bibinfo{volume}{32}, \bibinfo{publisher}{Curran Associates,
  Inc.}, \bibinfo{year}{2019}, pp. \bibinfo{pages}{7825--7835}.
%Type = Inproceedings
\bibitem[{Suzuki et~al.(2020)Suzuki, Takeno, Tamura, Shitara, and
  Karasuyama}]{pmlr-v119-suzuki20a}
\bibinfo{author}{S.~Suzuki}, \bibinfo{author}{S.~Takeno},
  \bibinfo{author}{T.~Tamura}, \bibinfo{author}{K.~Shitara},
  \bibinfo{author}{M.~Karasuyama},
\newblock \bibinfo{title}{{{Multi-objective {B}ayesian Optimization using
  Pareto-frontier Entropy}}},
\newblock in: \bibinfo{editor}{H.~D. III}, \bibinfo{editor}{A.~Singh} (Eds.),
  \bibinfo{booktitle}{Proceedings of the 37th International Conference on
  Machine Learning}, volume \bibinfo{volume}{119} of
  \textit{\bibinfo{series}{Proceedings of Machine Learning Research}},
  \bibinfo{publisher}{PMLR}, \bibinfo{year}{2020}, pp.
  \bibinfo{pages}{9279--9288}.
%Type = Article
\bibitem[{Forrester et~al.(2007)Forrester, Sóbester, and
  Keane}]{Forrester2007}
\bibinfo{author}{A.~I.~J. Forrester}, \bibinfo{author}{A.~Sóbester},
  \bibinfo{author}{A.~J. Keane},
\newblock \bibinfo{title}{{{Multi-fidelity optimization via surrogate
  modelling}}},
\newblock \bibinfo{journal}{Proc. R. Soc. A.} \bibinfo{volume}{463}
  (\bibinfo{year}{2007}) \bibinfo{pages}{3251--3269}.
%Type = Book
\bibitem[{Cover and Thomas(2012)}]{CT_book}
\bibinfo{author}{T.~M. Cover}, \bibinfo{author}{J.~A. Thomas},
  \bibinfo{title}{{{Elements of Information Theory}}},
  \bibinfo{publisher}{Wiley}, \bibinfo{edition}{2nd} edition,
  \bibinfo{year}{2012}.
%Type = Article
\bibitem[{Golovin and Krause(2011)}]{Golovin2011}
\bibinfo{author}{D.~Golovin}, \bibinfo{author}{A.~Krause},
\newblock \bibinfo{title}{{{Adaptive Submodularity: Theory and Applications in
  Active Learning and Stochastic Optimization}}},
\newblock \bibinfo{journal}{Journal of Artificial Intelligence Research}
  \bibinfo{volume}{42} (\bibinfo{year}{2011}) \bibinfo{pages}{427--486}.
%Type = Inproceedings
\bibitem[{Chen et~al.(2015)Chen, Hassani, Karbasi, and Krause}]{Chen2015}
\bibinfo{author}{Y.~Chen}, \bibinfo{author}{S.~H. Hassani},
  \bibinfo{author}{A.~Karbasi}, \bibinfo{author}{A.~Krause},
\newblock \bibinfo{title}{{{Sequential Information Maximization: When is Greedy
  Near-optimal?}}},
\newblock in: \bibinfo{booktitle}{Proceedings of The 28th Conference on
  Learning Theory}, volume~\bibinfo{volume}{40} of
  \textit{\bibinfo{series}{PMLR}}, pp. \bibinfo{pages}{338--363}.
%Type = Article
\bibitem[{L{{\'a}}zaro-Gredilla et~al.(2010)L{{\'a}}zaro-Gredilla,
  Qui{{\~n}}nero-Candela, Rasmussen, and Figueiras-Vidal}]{SSGP2010}
\bibinfo{author}{M.~L{{\'a}}zaro-Gredilla},
  \bibinfo{author}{J.~Qui{{\~n}}nero-Candela}, \bibinfo{author}{C.~E.
  Rasmussen}, \bibinfo{author}{A.~R. Figueiras-Vidal},
\newblock \bibinfo{title}{{{Sparse Spectrum Gaussian Process Regression}}},
\newblock \bibinfo{journal}{Journal of Machine Learning Research}
  \bibinfo{volume}{11} (\bibinfo{year}{2010}) \bibinfo{pages}{1865--1881}.
%Type = Article
\bibitem[{Pedregosa et~al.(2011)Pedregosa, Varoquaux, Gramfort, Michel,
  Thirion, Grisel, Blondel, Prettenhofer, Weiss, Dubourg, Vanderplas, Passos,
  Cournapeau, Brucher, Perrot, and Duchesnay}]{scikit-learn}
\bibinfo{author}{F.~Pedregosa}, \bibinfo{author}{G.~Varoquaux},
  \bibinfo{author}{A.~Gramfort}, \bibinfo{author}{V.~Michel},
  \bibinfo{author}{B.~Thirion}, \bibinfo{author}{O.~Grisel},
  \bibinfo{author}{M.~Blondel}, \bibinfo{author}{P.~Prettenhofer},
  \bibinfo{author}{R.~Weiss}, \bibinfo{author}{V.~Dubourg},
  \bibinfo{author}{J.~Vanderplas}, \bibinfo{author}{A.~Passos},
  \bibinfo{author}{D.~Cournapeau}, \bibinfo{author}{M.~Brucher},
  \bibinfo{author}{M.~Perrot}, \bibinfo{author}{E.~Duchesnay},
\newblock \bibinfo{title}{{{Scikit-learn: Machine Learning in {P}ython}}},
\newblock \bibinfo{journal}{Journal of Machine Learning Research}
  \bibinfo{volume}{12} (\bibinfo{year}{2011}) \bibinfo{pages}{2825--2830}.
%Type = Misc
\bibitem[{{{\relax U.S.~Geological Survey (USGS)}}()}]{USGS}
\bibinfo{author}{{{\relax U.S.~Geological Survey (USGS)}}},
  \bibinfo{title}{\url{https://pubs.usgs.gov/ds/121/grand/grand.html}},
  \bibinfo{year}{{}}.
%Type = Article
\bibitem[{Bojanov and Petrova(1998)}]{Bojanov1998}
\bibinfo{author}{B.~Bojanov}, \bibinfo{author}{G.~Petrova},
\newblock \bibinfo{title}{{{Numerical integration over a disc. A new Gaussian
  quadrature formula}}},
\newblock \bibinfo{journal}{Numer. Math.} \bibinfo{volume}{80}
  (\bibinfo{year}{1998}) \bibinfo{pages}{39--59}.
%Type = Incollection
\bibitem[{Brent(1973)}]{Brent1973}
\bibinfo{author}{R.~P. Brent},
\newblock \bibinfo{title}{{{Chapter 4: An Algorithm with Guaranteed Convergence
  for Finding a Zero of a Function}}},
\newblock in: \bibinfo{booktitle}{Algorithms for Minimization without
  Derivatives}, \bibinfo{publisher}{Prentice-Hall}, \bibinfo{year}{1973}.
%Type = Article
\bibitem[{{Virtanen et al.}(2020)}]{scipy}
\bibinfo{author}{P.~{Virtanen et al.}},
\newblock \bibinfo{title}{{{SciPy} 1.0: Fundamental Algorithms for Scientific
  Computing in Python}},
\newblock \bibinfo{journal}{Nature Methods} \bibinfo{volume}{17}
  (\bibinfo{year}{2020}) \bibinfo{pages}{261--272}.
%Type = Article
\bibitem[{Wilcoxon(1945)}]{Wilcoxon}
\bibinfo{author}{F.~Wilcoxon},
\newblock \bibinfo{title}{{{Individual Comparisons by Ranking Methods}}},
\newblock \bibinfo{journal}{Biometrics Bulletin} \bibinfo{volume}{1}
  (\bibinfo{year}{1945}) \bibinfo{pages}{80--83}.
%Type = Inproceedings
\bibitem[{Snoek et~al.(2012)Snoek, Larochelle, and Adams}]{Snoek2012}
\bibinfo{author}{J.~Snoek}, \bibinfo{author}{H.~Larochelle},
  \bibinfo{author}{R.~P. Adams},
\newblock \bibinfo{title}{{{Practical Bayesian Optimization of Machine Learning
  Algorithms}}},
\newblock in: \bibinfo{booktitle}{Advances in Neural Information Processing
  Systems}, volume~\bibinfo{volume}{25}, pp. \bibinfo{pages}{2951--2959}.
%Type = Inproceedings
\bibitem[{Akiba et~al.(2019)Akiba, Sano, Yanase, Ohta, and Koyama}]{Optuna}
\bibinfo{author}{T.~Akiba}, \bibinfo{author}{S.~Sano},
  \bibinfo{author}{T.~Yanase}, \bibinfo{author}{T.~Ohta},
  \bibinfo{author}{M.~Koyama},
\newblock \bibinfo{title}{{{Optuna: A Next-generation Hyperparameter
  Optimization Framework}}},
\newblock in: \bibinfo{booktitle}{Proceedings of the 25rd {ACM} {SIGKDD}
  International Conference on Knowledge Discovery and Data Mining}.
%Type = Unpublished
\bibitem[{Ramos-Carreño(2020)}]{dcor2020}
\bibinfo{author}{C.~Ramos-Carreño}, \bibinfo{title}{dcor},
  \bibinfo{year}{2020}. \bibinfo{note}{\url{https://github.com/vnmabus/dcor}}.
%Type = Article
\bibitem[{Ramos-Carreño and Torrecilla(2023)}]{RAMOSCARRENO2023101326}
\bibinfo{author}{C.~Ramos-Carreño}, \bibinfo{author}{J.~L. Torrecilla},
\newblock \bibinfo{title}{{dcor: Distance correlation and energy statistics in
  Python}},
\newblock \bibinfo{journal}{SoftwareX} \bibinfo{volume}{22}
  (\bibinfo{year}{2023}) \bibinfo{pages}{101326}.

\end{thebibliography}
\end{document}